\documentclass[10pt, journal, compsoc, letterpaper]{IEEEtran}

\usepackage[T1]{fontenc}
\usepackage[nocompress]{cite}
\usepackage{hyperref}
\hypersetup{colorlinks=false}
\ifCLASSINFOpdf
\usepackage[pdftex]{graphicx}
\graphicspath{{pics/}}
\DeclareGraphicsExtensions{.pdf,.jpg,.png}
\else
\fi
\usepackage{enumitem}
\setlist[description]{leftmargin=*, align=left, itemsep=0.5em}

\usepackage{amsmath}
\usepackage{amssymb}
\newcommand{\argmin}[1]{\ensuremath{\underset{#1}{\text{argmin}}}}
\newcommand{\argmax}[1]{\ensuremath{\underset{#1}{\text{argmax}}}}

\renewcommand{\vec}[1]{\mathbf{#1}}

\usepackage{bbm}
\usepackage[T1]{fontenc}

\renewcommand{\arraystretch}{1.6}

\ifCLASSOPTIONcompsoc
  \usepackage[caption=false,font=footnotesize,labelfont=sf,textfont=sf]{subfig}
\else
  \usepackage[caption=false,font=footnotesize]{subfig}
\fi
\begin{document}
\title{An In-Depth Analysis of Visual Tracking with Siamese Neural Networks}
\author{Roman~Pflugfelder,~\IEEEmembership{Member,~IEEE}
\IEEEcompsocitemizethanks{\IEEEcompsocthanksitem R. Pflugfelder is with the Centre of Digital Safety \& Security, AIT Austrian Institute of Technology and with the Institute of Visual Computing \& Human-Centred Technology, University of Technology Vienna.\protect\\
E-mail: roman.pflugfelder@\{ait.ac.at|tuwien.ac.at\}
}}
%\IEEEspecialpapernotice{(Draft Article)}
\IEEEtitleabstractindextext{%
\begin{abstract}
This survey presents a deep analysis of the learning and inference capabilities in nine popular trackers. It is neither intended to study the whole literature nor is it an attempt to review all kinds of neural networks proposed for visual tracking. We focus instead on Siamese neural networks which are a promising starting point for studying the challenging problem of tracking. These networks integrate efficiently feature learning and the temporal matching and have so far shown state-of-the-art performance. In particular, the branches of Siamese networks, their layers connecting these branches, specific aspects of training and the embedding of these networks into the tracker are highlighted. Quantitative results from existing papers are compared with the conclusion that the current evaluation methodology shows problems with the reproducibility and the comparability of results. The paper proposes a novel Lisp-like formalism for a better comparison of trackers. This assumes a certain functional design and functional decomposition of trackers. The paper tries to give foundation for tracker design by a formulation of the problem based on the theory of machine learning and by the interpretation of a tracker as a decision function. The work concludes with promising lines of research and suggests future work.

\end{abstract}
% Note that keywords are not normally used for peer-review papers.
}
\maketitle
\IEEEdisplaynontitleabstractindextext
\IEEEpeerreviewmaketitle

%%%%%%%%%%%%%%%%%%%%%%%%%%%%%%%%%%%%%%%%%%%%%%%%%%%
\IEEEraisesectionheading{
\section{Introduction} \label{sec:introduction}}
% tracking is a challenge; quantitative analysis show minor effects
\IEEEPARstart{V}{isual} tracking has received significant attention over the last decades and it is still a very active research field. This is well documented by hundreds of publications each year and several community activities\footnote{\url{http://videonet.team} summarises these activities; 24/07/2018.}. Numerous applications in diverse application fields trigger these activities such as in robotics~\cite{held-RSS2016}, augmented reality~\cite{zhang-CVPR2015}, solar forecasting~\cite{chow-solar2015}, microscopy~\cite{maska-PMC2014}, biology~\cite{bozek-CVPR2018} and surveillance~\cite{jiang-AVSS2017}. We ephasise that this personal choice of samples reflects only a tiny number of potential application fields.

The lack of quantitative comparison makes common findings in different trackers and theoretical insights a very hard task.
Rather less is currently known about the tracking problem, i.e. the tracking of arbitrary objects in arbitrary scenes. Since 2013, new datasets and methodologies have improved this situation~\cite{wu-CVPR2013, smeulders-TPAMI2013, kristan-TPAMI2016,  liang-TIP2015, mueller-CORR2018, valmadre-CORR2018, mueller-ECCV2016, liang-ICRA2018}, but still after five years of evaluation, progress in the understanding of tracking is encouraging and sobering at the same time.

% main results of this quantitative analysis; improvement but no breakthrough (deep learning)
One challenge of tracking is that a tracker needs to address accuracy in object localisation, robustness to visual nuisances and computational efficiency all at once. The VOT (Visual Object Tracking) challenges\footnote{\url{http://www.votchallenge.net}; 24/07/2018} show that advanced correlation filters, convolutional features and their seamless integration for end-to-end training bring significant improvement to these aspects when compared to the 2013 online learning state of the art. The AuC (Area Under the Curve) of success rate vs. overlap for the OTB-2013 (Object Tracking Benchmark) OPE experiment\footnote{The OPE (One Pass Experiment) initialises the tracker and does not interact for the whole sequence of test images.} improved by 31.65\,\% from 47.3\,\% (Struck - structured SVM~\cite{hare-ICCV2011}) to 69.2\,\% (C-COT\_CFCF~\cite{kristan-VOT2017, gundogdu-TIP2018})\footnote{MDNet~\cite{nam-CVPR2016} reports so far to hit on OTB-2013 the 70\,\% with 70.8\,\%, but the tracker is computationally very demanding (1\,fps on a GPU). Guo et al.~\cite{guo-ICCV2017} observe unfair training on the test set. By training on ILSVRC-2015 they report for MDNet 61.9\,\%.}. The raw VOT-2017 R-scores show even a drop by 78.33\,\% from 1.297 to 0.281 failures per 100 frames. However, these relative quantitative improvements might lead us astray, as average failure in terms of absolute numbers is with 4.2 failures per 1\,min 25\,fps video still very large. Similarly, the raw VOT-2017 A-scores show just a moderate increase by 21.77\,\% from 0.418 to 0.509\,IoU (Intersection of Union), i.e. on average the trackers' bounding boxes are 50\,\% congruent with the ground truth which is definitely not satisfactorily. A closer look to R-scores show correlation filters brought most of the improvement (23.12\,\%) contrary to deep learning (8.53\,\%) which shows that until now, neural networks have not brought the breakthrough in tracking as it brought in categorial object recognition (ILSVRC-2010~\cite{russakovsky-IJCV2015}).

% empirical evaluation might not give the important insights for a breakthrough
What we further observe is a saturation in A-score and R-score in small intervals below 0.5\,\% and 0.7\,\% respectively and regardless of the approach. Larger and more diverse datasets~\cite{mueller-CORR2018, valmadre-CORR2018} might help the empirical evaluation in future to identify the promising approaches. But it is unclear if these clear drops in the scores for the promising approaches will occur as tracking is complex and so many different approaches with indistinguishable results exist. We believe that discrimination alone will not give significant insights in tracking and that empirical evaluation goes hand in hand with a throughout qualitative analysis.

% qualitative analysis marked important developments but does not affirm conclusions
Comparative paper studies of tracking have been published~\cite{moeslund-CVIU2006, yilmaz-ACM2006, cannons-TR2008, yang-Neurocomputing2011, li-ACM2013}. Yilmaz et al.~\cite{yilmaz-ACM2006} and Cannons~\cite{cannons-TR2008} propose comprehensive taxonomies by identifying building blocks such as representation\footnote{This paper understands representation as an implementation of a description, i.e. concrete data structures and feature transforms.}, initialisation, prediction, association and adaptation and discuss pros and cons of features such as points, edges, contours, regions and their combinations. Li et al.~\cite{li-ACM2013} analyse only the description, with a comprehensive summary of global and local features and a review on generative, discriminative and hybrid learning approaches. Yang et al.~\cite{yang-Neurocomputing2011} do similar by identifying global and local features  and their integration with object context by using particle filters as inference framework. They discuss online learning in this framework by summarising generative, discriminative and combined methods. Although these descriptive studies were able to mark important developments in research, it becomes impossible to enumerate all approaches, even  in recent literature and to recognise the tiny but important design aspects and their differences and consequences.
%\footnote{Research is very active since 2017. For example, Valmadre et al.~\cite{valmadre-CVPR2017} received 56 citations (Google Scholar; 05/30/2018). Tao et al.~\cite{tao-CVPR2016} received 121 citations, 8 in 2016, 75 in 2017.}.

% qualitative analysis based on quantitative analysis
Therefore, comparative studies based on empirical evaluation have recently received some attention~\cite{smeulders-TPAMI2013, liang-TIP2015, li-PR2018}. Smeulders et al.~\cite{smeulders-TPAMI2013} compare 19 trackers by their predicted bounding boxes on the ALOV (Amsterdam Library of Ordinary Videos) dataset. The trackers were chosen by their online learning capability. Sparse and local features seem appropriate to handle rapid changes such as occlusion, clutter and illumination. There is evidence that complex descriptions are inferior to simpler ones. The experiments show little conceptual consensus among the trackers and find none of the trackers superior.
Liang et al.~\cite{liang-TIP2015} created a new TColor-128 dataset and compare 25 trackers on 128 color image sequences. 16 grayscale trackers are extended by ten different features of color. Color is an important cue and different trackers show preference to specific color models. Some extended grayscale trackers even outperform specific color trackers. Color helps with deformation, rotation and illumination changes, while occlusion and motion blur still remain challenges for color. 
Li et al.~\cite{li-PR2018} conduct experiments with 16 trackers, all of them chosen by their deep learning capabilities. They build a taxonomy of deep trackers, evaluate their performance on OTB, VOT and assess the approaches based on the qualitative and quantitative analysis.

% aim of the work
The aim of this work is to take the same combined comparative and empirical approach but instead of a broad comparison of diverse tracking approaches we focus our analysis on a small number of trackers all belonging to a specific class. This allows a comparative analysis of the approach where even little details are considered. We believe that such in-depth analysis is as valuable to gain substantial insight into tracking as a broad comparison.

% why Siamese networks?
The class of trackers we choose are trackers based on Siamese neural networks. These networks are a good starting point for a deep analysis, as they are the simplest networks applicable for tracking. Learning and inference with a Siamese network~\cite{baldi-JNC1993, bromley-NIPS1993} is still very general and a promising approach for many problems such as face verification and recognition~\cite{chopra-CVPR2005, taigman-CVPR2014, parkhi-BMVC2015, schroff-CVPR2015}, areal-to-ground image matching~\cite{lin-CVPR2015}, stereo matching~\cite{zbontar-CVPR2015}, patch matching~\cite{han-CVPR2015, simo-serra-ICCV2015, zagoruyko-CVPR2015}, optical flow~\cite{dosovitskiy-ICCV2015}, large-scale video classification~\cite{karpathy-CVPR2014} and one-shot character recognition~\cite{koch-ICMLW2015}. We believe that a throughout understanding of tracking using Siamese networks will lead to important new general insights.

A Siamese network is a Y-shaped neural network that joins two network branches to produce a single output. The idea originated 1993 in fingerprint recognition~\cite{baldi-JNC1993} and signature verification~\cite{bromley-NIPS1993}, where the task is to compare two imaged fingerprints or two hand-written signatures and to infer identity. A Siamese network captures the comparison of the preprocessed input as a function of similarity, more precisely a function of the class of Lipschitz functions~\cite{shalev-shwartz-BOOK2014} $f: [-1,1]^d \to [-1,1]$, with the advantageous ability to learn similarity and the features jointly and directly from the data. In statistical decision theory~\cite{berger-BOOK1985} and machine learning~\cite{shalev-shwartz-BOOK2014}, such function of similarity is understood as decision function and feature learning is known as experimental design. Despite the generality and usefulness of Siamese networks in various applications, relatively less is known about their statistical foundation and properties~\cite{blackwell-SMSP1951, nguyen-AOS2009}. 

% why neural networks for tracking?
In principle, neural networks have the ability to combine multiple tasks~\cite{doersch-ICCV2017, kalogeiton-ICCV2017}, even heterogeneous models~\cite{ranftl-DAGM2014, knoebelreiter-CVPR2017} and allow end-to-end training with labelled data by using the idea of back-propagation~\cite{rumelhart-NATURE1986}. This property of neural networks allows a formulation of multi-task learning and inference which is very advantageous against other approaches, because, as Yilmaz et al.~\cite{yilmaz-ACM2006} and Cannons~\cite{cannons-TR2008} have noted,  tracking itself is a combination of multiple building blocks~\cite{cannons-TR2008}. Neural networks, furthermore, show great success in other fields where data is also spatiotemporal as in tracking such as in speech recognition~\cite{graves-ICML2006} and lip reading~\cite{assael-CORR2016}.
Siamese networks are a promising approach for a deeper analysis and subsequent discussion of tracking as they realise conceptually the simplest way of visual description and temporal matching. They are trivial recurrent networks, thus a good starting point to study initialisation, prediction and adaptation in tracking. Siamese networks are computationally very efficient in both inference and in learning and have so far shown state-of-the-art performance in accuracy and robustness.

% seeing the tracking from a machine learning point of view
Instead of formulating tracking ad hoc as a series of processing steps or for example as filtering problem in the framework of Bayesian analysis~\cite{challa-book2011}, this work suggests to interpret the tracking problem as decision function which changes the view of tracking from a classical inference to a data-driven inference and learning approach. Learning tracking then can be naturally understood and formulated by applying machine learning and statistical decision theory respectively. We further analyse the design of the decision function by de-composing the function into functions of pre- and post-processing as well as the neural network. We use for this analysis a Lisp-like formalism for describing the decision function which is an alternative to the comparison of trackers by using graphical illustrations.

% Quantitative analysis
An important result of our study is the understanding that current methodology used in scientific work to generate quantitative results is often inappropriate. Results published in the past papers are frequently irreproducible and variant to the chosen training data. Test data is constantly growing which influences negatively the comparability of results over longer time.
It is no aim of this work to conduct new empirical experiments or introduce new evaluation methodologies. This paper's quantitative analysis is based on the evaluation results of available benchmarks and papers.

% contribution of this work
To summarise, the contribution of this work are new insights into the use of Siamese networks for learning tracking and tracker inference by: 
\begin{enumerate}
\item a new formulation of tracking inference and learning from the viewpoint of stochastic processes, statistical decision theory and machine learning,
\item seeing the design of trackers as functional composition of the decision function which is elegantly applicable to neural networks,
\item describing these functional compositions with a novel prefix Lisp-like formalism which allows a new way to compare trackers and their details,
\item and finally an in-depth analysis and detailed comparison of nine recent trackers that use Siamese networks.
\end{enumerate}

% explain how the paper is organised
The work is organised as follows: Section~\ref{sec:background} formulates the problem of tracking and tracker design by considering the traditional and the connectionist view, i.e. for short a view where we use neural networks for tracking. Current challenges of tracking are stated. Section~\ref{sec:analysis} gives the analysis of recent trackers where emphasis is on the network architecture, the network input and output and its interpretation, the aspects of training and finally the embedding of the network for inference in the overall tracking method. Section~\ref{sec:discussion} then discusses the differences in the network branches, the connection of branches, the training and the trackers themselves, to offer some insights into the use of Siamese networks for learning tracking and tracker inference. Section~\ref{sec:conclusion} concludes our study and gives an overview of potential future work based on a reflection of the paper's results w.r.t. the aforementioned tracking challenges.

%%%%%%%%%%%%%%%%%%%%%%%%%%%%%%%%%%%%%%%%%%%%%%%%%%%
\section{Background}\label{sec:background}
% visual tracking as important problem in science
%Awareness of objects in the environment is fundamental for life~\cite{plessner-book1975}. Visual perception is essential to build high-level cognitive abilities which is the reason why the visual perceptual system takes such a large proportion of the human cortex~\cite{kandel-book2000}. As for humans and nearly all animals, visual sensory and tracking is a foundational perceptual ability for the development of high-level cognitive capabilities. Human vision and awareness of things is certainly a process in space and time as most of the critical information for human decisions emerges directly by the spatiotemporal stimuli in front of us. Physical objects or phenomena in the scene cause usually these stimuli. While some cognitive psychologists study this process by a descriptive characterisation based on the stimuli and their consequences~\cite{grossberg-TCS2000, valuch-JV2017}, others directly state the process as analytical mathematical problem~\cite{marcus-BOOK2003}.

% visual tracking in engineerig
Tracking in econometrics~\cite{durbin-BOOK2012}, control theory~\cite{challa-book2011, li-TAES2003}, radar~\cite{barshalom-book2001, blackman-BOOK1999} and computer vision~\cite{maggio-book2011, betke-book2016} is the stochastic problem of estimating and predicting random variables such as object appearance, position, dynamics and behaviour by the way of collecting sufficient evidence from usually multiple sensory inputs such as a sequence of camera images. The process is assumed probabilistic as many factors of the problem formulation are typically uncertain or effectively unknown. For example, correspondence of pixels in the sequence is usually unknown. A general analytical formulation becomes quickly statistically and computationally complex or even intractable, therefore a priori knowledge of the application by additional assumptions is frequently needed. A good model of object and tracker capturing the constraints and assumptions of the application greatly outperforms very general trackers for arbitrary objects~\cite{li-TAES2003}.

It is important to note that some of the variables vary over time. For example in visual tracking, the position or occupied region of an object in the video changes with the object's movement. Tracking differs in this respect from other inference problems assuming stationary processes and combines statistical decision theory~\cite{berger-BOOK1985} with stochastic process theory and related theories such as time series analysis. Many classical solutions build on state-space models and Bayesian Analysis~\cite{challa-book2011} and combinatorics such as Multiple Hypothesis Tracking (MHT)~\cite{blackman-AESM2004} for multiple targets.

%%%%%%%%%%%%%%%%%%%%%%%%%%%%%%%%%%%%%%%%%%%%%%%%%%%
\subsection{Tracking Problem} \label{sec:problem}
% show an illustration of the tracking problem (inference)
\begin{figure}
\centering
\includegraphics[width=.6\columnwidth]{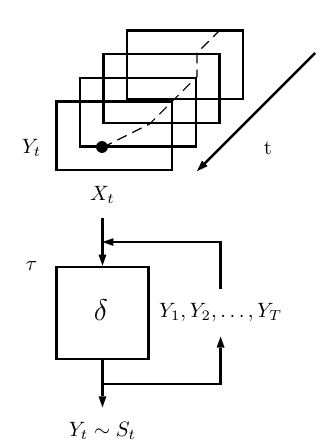}
\caption{Illustration of the tracking problem.}
\label{fig:problem}
\end{figure}

% define tracking inference as a stochastic process
As tracking is a problem of inference (Fig.~\ref{fig:problem}), it is in its roots a stochastic process that is described by a set of random variables\footnote{Variables are named states in state-space models and hidden or latent variables in graphical models.}
\begin{equation}
Y(t) = \{ Y_t \in \mathcal{M} \times \mathcal{A}, t \in \mathbb{N} \}.
\label{eq:Y}
\end{equation}
The variables $Y_t$ capture measurements in a combined motion space $\mathcal{M}$ and appearance space $\mathcal{A}$ usually in discrete time steps $t \in \{1,2,\dots\}$. For example, $Y_t$ might describe the position and object's dynamics in 3-D space, but in computer vision very often,  $Y_t$ describes the well known bounding box in the image plane together with its enclosed image template or patch. The process is stochastic, as a precise mathematical description of motion and images is possible~\cite{barshalom-book2001} but in the most cases computationally intractable. For example, think of the complex motion of individual ants in ant colonies~\cite{khan-TPAMI2006} or the Brownian motion of interacting water molecules in clouds~\cite{chow-solar2015}.

% define the tracker function
The stochastic process $Y(t)$ is realised by a generator function $\tau : \mathbb{N} \to \mathcal{M} \times \mathcal{A}, t \to \tau(t)$, which we will call the tracker function or the tracker. One way to derive a tracker function is to see the problem of inference from the view of metrology. This yields the famous measurement equation~\cite{schruefer-BOOK2012}
\begin{equation}
\tau(t) = S_t + \epsilon_\tau + N_t
\label{eq:tau}
\end{equation}
which relates $\tau(t)$ with the unknown state of nature $S_t$ and errors $\epsilon_\tau$ and $N_t$ respectively. $S_t$ and the underlying in most cases non-stationary distribution of $Y(t)$ is usually unknown which makes tracking a very challenging task.
Measurements done by $\tau(t)$ and the true values differ either by errors $\epsilon_\tau$ made systematically by the tracker or by random errors $N_t$ caused by the limitations of modelling $\tau(t)$. The propagation of systematic errors is a serious challenge in tracking, because it introduces drift over time which is caused by a miss-match of object and its artificial description. The initial state $S_1$ is usually assumed to be known, hence the initial element $Y(1) = \{ Y_1 \}$ is also given by $\tau(1) = S_1$. 

% tracker function of a camera
A combined video acquisition and tracking system realises $\tau(t)$ basically in two steps. It acquires a sequence of images
\begin{equation}
X(t) = \{ X_t \in \mathcal{I}_C, t \in \mathbb{N} \}
\end{equation}
in space $\mathcal{I}_C$, followed by an analysis that is expressible by a function
\begin{equation}
\delta :  \mathcal{I}^t_C \to \mathcal{M} \times \mathcal{A}, X(t) \to \delta ( X(t) ).
\label{eq:delta}
\end{equation}
Further assumptions about the object, the scene and the camera might be made by $\delta(\cdot)$. We note that $\delta(\cdot)$ might in principle have different interpretations. Beside metrology (Equation~\ref{eq:tau}), in time-series analysis, $\delta(\cdot)$ is a state space model, in statistical decision theory and machine learning it is a decision function. In this latter interpretation, the tracker is basically learnable from sample data with given but unknown distribution. Learning tracking is not as easy and naturally understandable with the aforementioned other theories.
%It is important to note that the camera and the analysis are not independent choices. Camera and analysis give a combined measurement method. In practice, a particular camera might simplify the analysis considerably. For example, consider tracking the 3-D pose of an object then a Time-of-Flight (ToF) camera that measures the depth to the object directly offers great advantages against a classical camera.

% introduce decision function
Under this consideration, the final tracker function for cameras is
\begin{equation}
\tau(t) = \delta (X(t))
\label{eq:tau-1}
\end{equation}
which we name Type-1 tracking. The main characteristic of Type-1 trackers is that they exploit the sequence of images without considering previous measurements. Such trackers are conservative as they rely fully on the prior information given by $S_1$. They are therefore less prone to systematic errors, to error propagation and finally to concept drift, e.g. in case the object's description is part of the measurements $Y(t)$. Type-1 tracking is well known in literature as tracking-by-detection~\cite{grabner-CVPR2006, avidan-CVPR2005}.

% Type-1 and Type-2
However, most trackers are not of Type-1. Usually successive measurements correlate as the object undergoes a continuous motion in 3-D space and the acquisition of $X(t)$ is fast enough. These important assumptions yield Type-2 tracking
 \begin{equation}
 \tau (t) = \delta( X(t), Y(t-1) ),
 \label{eq:tau-2}
\end{equation}
where previous measurements are considered by $\tau(t)$.

Equation~\ref{eq:tau-2} is contrary to Equation~\ref{eq:tau-1} a difference equation which expresses the fundamental recurrent nature of tracking. All trackers using a Bayesian analysis such as filtering methods fall under Type-2. Equation~\ref{eq:tau-2} assumes theoretically unlimited resources as it accesses all images and previous measurements. Practical trackers are derived by introducing a parameter $n > 1$ that limits the tracker's history which gives
\begin{equation}
\tau (t) = \delta( X(t)\backslash X(t-n), Y(t-1)\backslash Y(t-n) ).
\label{eq:tau-n}
\end{equation}
For example, $n = 2$ yields the important class of Markov trackers $\tau(t) = \delta( \{ X_t, X_{t-1} \}, Y_{t-1} )$ following the Markov assumption (Type-2M). For $n > 2$ the tracker considers a finite $n$-batch of history data. Type-2 tracking is inherently prone to error propagation, because of the systematic error $\epsilon_\tau$ in Equation~\ref{eq:tau}.

%%%%%%%%%%%%%%%%%%%%%%%%%%%%%%%%%%%%%%%%%%%%%%%%%%%
\subsection{Tracker Design}
% derive the design space
Statistical decision theory and machine learning let us design a decision function for tracking in many ways. We saw in Section~\ref{sec:problem} that a proper decision function infers measurements of the object from a sequence of images, where $\delta(\cdot)$ is an unbounded operator~(Equation~\ref{eq:delta}). We will restrict $\delta(\cdot)$ in this paper to compositions of two partial tracker functions $\delta_0(\cdot)$ and $\delta_1(\cdot)$, including a functional $\sigma \in \Sigma_S^i$ representing convolutional networks from five classes ($1\le i \le 5$) of the Siamese neural network $\Sigma_S^i$ (Fig.~\ref{fig:net}), i.e.
 \begin{eqnarray}
 \delta ( X(t), Y(t-1) ) & = & \Delta_{X(t), Y(t-1)}(\sigma)
\end{eqnarray}
with 
 \begin{eqnarray} 
\Delta_{X(t), Y(t-1)}(\sigma)  & = & \delta_1(\sigma, \delta_0 ( X(t), Y(t-1) ) ).
 \label{eq:sigma}
\end{eqnarray}

$\Delta_{X(t), Y(t-1)}(\cdot)$ needs here to be more expressive thus more complex than the straight forward composition
\begin{eqnarray}
 \delta ( X(t), Y(t-1) ) & = & \delta_1 \circ \sigma \circ \delta_0 ( X(t), Y(t-1)  )
\end{eqnarray}
to be able to capture the studied trackers which this paper will compare (Section~\ref{sec:analysis}). We note that the design space of $\delta(\cdot)$ is certainly much larger. Less is known about this design space and relatively less has been done to characterise this space, except for the case the decision method is derived from a probability model and from the view of Bayesian analysis~\cite{challa-book2011}.

%%%%%%%%%%%%%%%%%%%%%%%%%%%%%%%%%%%%%%%%%%%%%%%%%%%
\subsubsection{A Traditional View}\label{sec:traditional}
% show illustration of the building blocks
\begin{figure}
\begin{center}
\includegraphics[width=\columnwidth]{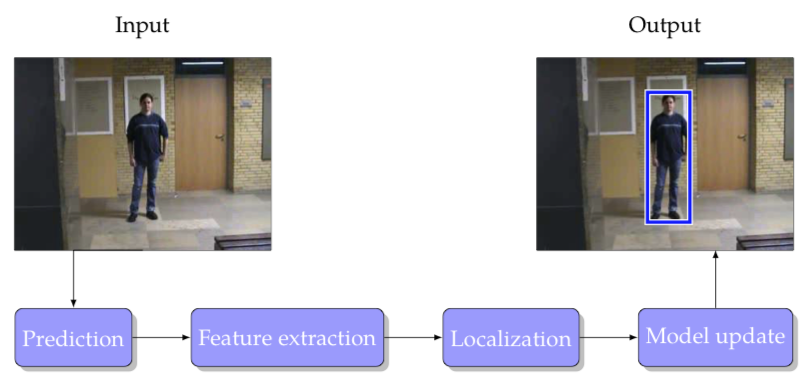}
\caption{Building Blocks of Traditional Tracker Design.}
\label{fig:building-blocks}
\end{center}
\end{figure}

% explanation
Traditional tracker design constitutes building blocks~\cite{wang-ICCV2015, nebehay-PHD2016} (Fig.~\ref{fig:building-blocks}). Given a motion model of the object, the tracker's predictor infers first intermediate measurements. These predictions restrict then a search region where features are extracted and compared against the currently existing description of the object (localisation). This description is then either adapted with the new features by a model update or kept fixed, depending on the design of $Y_t$. The updater and the final post-processing estimate the new state $Y(t)$ based on the prediction and the measurements taken.

% learning
These building blocks are usually seen as fixed such as in the Kalman filter~\cite{kalman-TASME1960} or Mean-shift tracker~\cite{comaniciu-TPAMI2002}. Section~\ref{sec:problem} introduces the tracker as (partially) unknown decision function, e.g. a $\theta$-parametrised family of functions. Learning tracking can then be understood as the procedure to minimise the empirical risk~\cite{shalev-shwartz-BOOK2014}
\begin{equation}
R_{\delta, X(t), Y(t)} ( \theta ) = \mathbb{E}_{X(t), Y(t)} [ l(\delta_\theta( X(t),Y(t-1) ),Y_t) ].
\label{eq:risk}
\end{equation}

$l(\cdot)$ is a loss function that measures the error made by the decision function to given spatiotemporal sample data <$X(t)$, $Y(t)$>. The decision function is then given by
\begin{equation}
\theta^*_{X(t), Y(t)} = \argmin{\theta}~R_{\delta, X(t), Y(t)} ( \theta ).
\label{eq:Rmin}
\end{equation}

% uncertainty
As $Y(t)$ is a stochastic process, uncertainty might be modelled by the moments~\cite{kalman-TASME1960} or probability density function~\cite{isard-IJCV1998}. Bayesian analysis of $\delta(\cdot)$ is then very popular~\cite{challa-book2011, barshalom-book2001, maggio-book2011} as we can derive $\delta(\cdot)$ easily from a probability model, e.g. Gaussian. Furthermore, the prior is in tracking well defined as long as the initial true measurement $S_1$ is available. In this case, the empirical risk is replaced by the Bayesian risk for learning.

%%%%%%%%%%%%%%%%%%%%%%%%%%%%%%%%%%%%%%%%%%%%%%%%%%%
\subsubsection{A Connectionist View}
% show illustration of the design of tracking inference from the view of connectionism
\begin{figure}
\begin{center}
\includegraphics[width=\columnwidth]{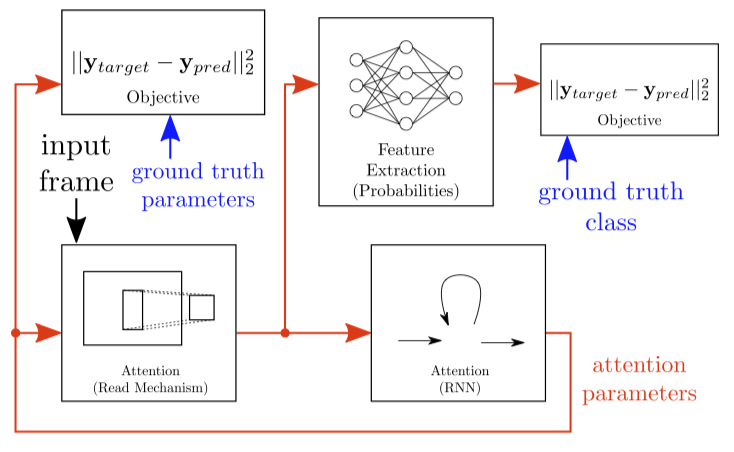}
\caption{An example of a Connectionist view of tracker design. Inference of the latent variables (measurements) $\vec{y}_\text{pred}$ is basically done by using the sequence of images in three complementary pathways: the who-pathway (feature extraction), the where-pathway (attention) and the why-pathways (objective) comparing the measurements with the ground truth $\vec{y}_\text{target}$. Image is from Kahou et al.~\cite{kahou-CVPRW2017}.}
\label{fig:pathways}
\end{center}
\end{figure}

% explanation
The re-discovery of neural networks in 2010 has brought again the connectionist view to tracking. Bazzani et al.~\cite{bazzani-ICML2011} designs the data flow of connectionist trackers by three fundamental pathways (Fig.~\ref{fig:pathways}). The who-pathway collects evidence of objects' appearance and shape as well as their identities, similar to the ventral pathway in the human brain~\cite{kandel-book2000}. Then, the where-pathway collects evidence of objects' location and motion (derivatives of location with respect to time) similar to the dorsal pathway in the human brain and finally the why-pathway that formalises the objectives of learning~\cite{kahou-CVPRW2017, kosiorek-NIPS2017}. This latter pathway represents meaning about the problem which is in this form not explicitly available in the traditional view.

% advantage
The building blocks as seen by the traditional view are captured by different network layers. These layers are interconnected and the links constitute these pathways. The complexity of the whole network and the choice of the network architecture has direct consequence to the robustness and accuracy of the tracking results as well as to the computational efficiency of the tracking algorithm. What is interesting is that $\delta(\cdot)$ might be approximated by a network of building blocks instead of the traditional pipeline and that $\delta(\cdot)$ is learnable form sample data in an end-to-end training manner. Layers or tasks might be very heterogeneous, i.e. might represent very different mathematical models, as long as they are differentiable. Beside the exceptional capability of networks to learn features, connectionism might offer a new design technique to better cope with the complexity of the tracking problem.

%%%%%%%%%%%%%%%%%%%%%%%%%%%%%%%%%%%%%%%%%%%%%%%%%%%
\subsubsection{Learning Principles}
Learning rigorously the functional parts of $\delta(\cdot)$ holistically by a network is still in its infancy. Equation~\ref{eq:Rmin} describes supervised learning which has been used for the tracking problem since 2015. It assumes large training data comprising sequences of images with labelled objects, e.g. bounding boxes which is for tracking very hard to achieve. Therefore, other learning principles are popular in tracking:

\begin{description}
\item[Unsupervised and semi-supervised learning] is a setting where the tracker adapts its description of the object to unlabelled video frames.  Semi-supervised learning differs as it exploits available spatiotemporal structure of the unlabelled video which poses additional constraints on the estimation of the parameters. This helps the learner to converge quicker to the best solution.
%For both principles adaptivity is important as trackers might be able to learn over their whole lifetime which poses the problem of limited computational resources and memories. This principle refers to learners that are able to forget the past when it is no longer relevant.

\item[Online learning] is the predominant way of learning tracking. Adaptivity is important as trackers might be able to learn over their whole lifetime. After the learning step, the data is usually replaced by new incoming image data. Early work targeted mainly on surveillance applications, therefore tracking was used to follow categorial objects such as vehicles and persons and learning tracking focused either on the categories by improving detectors or on instances by using generic local features.

\item[Offline learning] exploits initial given data, e.g. to learn a priori motion, however, learning tracking offline was underrated and has been rediscovered recently, e.g. to learn categorial properties or in the other extreme, to learn discriminative features among individuals such as for re-identification.
\end{description}

While online learning has been seen in an unsupervised or semi-supervised way, offline learning is usually supervised and limited by the availability of labelled data. Fortunately, new results suggest ways to overcome this problem~\cite{valmadre-CORR2018}. Although supervised learning gains  popularity, learning tracking from unlabelled or weakly labeled data is still the standard of learning tracking in literature.
%Human vision research shows that learning tracking is an intentional, preconscious process, so there are good arguments for learning tracking 

%%%%%%%%%%%%%%%%%%%%%%%%%%%%%%%%%%%%%%%%%%%%%%%%%%%
\subsection{Tracking Challenges}\label{sec:challenges}
%% illustrate these challenges
%\begin{figure}
%\begin{center}
%\includegraphics[width=\columnwidth]{challenges}
%\caption{Tracking challenges.}
%\label{fig:tracking-challenges}
%\end{center}
%\end{figure}

% summarise challenges based on survey literature
Tracking is in its generality unsolved, only partial solutions to Equation~\ref{eq:tau-n} under rather specific constraints are available in literature~\cite{moeslund-CVIU2006, yilmaz-ACM2006, cannons-TR2008, yang-Neurocomputing2011, li-ACM2013, smeulders-TPAMI2013, liang-TIP2015, li-PR2018}. These studies name a variety of problems, characterised by five important challenges subject of current research:
\begin{description}
\item[Complexity:] The higher the complexity, i.e. the information capacity of the description, the more information can be inferred by a set of latent variables. For example, the image template predicts either bounding boxes or pixelwise segmentations of the object but not 3-D pose which needs descriptions of depth. Higher complexity, however, comes with difficulty of collecting unambiguous evidence about the predictions of latent variables. Descriptions should be made simple to be informative but as complex to meet application specific demands. 

\item[Uncertainty:] Trackers conform to a specification of quality parameters during a period of time such as the robustness needed for tolerating changes and for preventing failure. The description of tracking needs to be invariant to temporal changes of the scene, objects, camera and their relationships. For example, changes in object appearance need to be tolerated by the tracker. At this point, we have to distinguish vagueness and uncertainty. While vagueness refers to controllable risks of specified changes, uncertainty refers to risks of possible changes, i.e. a potential subset of changes is unspecified hence unknown to the tracker.
%An example is a tracker for arbitrary objects. Such trackers become important. Descriptions of arbitrary objects are therefore needed. Such descriptions should be sufficiently made time-invariant to meet application specific demands.

\item[Initialisation:] Humans or object detectors initialise usually trackers in the first video frame. Unfortunately, object detectors are limited to certain object categories. Online learning a detector of the individual object is very important to re-initialise the tracker after full occlusions. Automatised initialisation of arbitrary objects is a huge challenge as initialisation assumes currently the human in the loop.
%In case of uncertainty, a generalised initialisation is needed which is able to handle arbitrary objects. Such an attentive tracker detects and tracks salient objects occurring in the video. For example, a salient object might be an image region of atypical spatiotemporal characteristics compared to its neighbourhood. Attentive mechanisms based on saliency are also found in human vision. Trackers for arbitrary objects should be made attentive to fully exploit their potential.

\item[Computability:] Efficiency of inference is highly important for the practical use of trackers. Many solutions turn inference very quickly to computational intractability. A balanced approach satisfying accuracy, robustness and speed is therefore still a significant challenge.
%has to be as efficient and accurate as possible, being able to compute predictions of latent variables and satisfying quality parameters at the same time. Complexity of the description and the prediction of the latent variables go hand in hand, turning  A principled approach balances both, leading to an optimal graceful degradation of inference.

\item[Comparison:]
Trackers need to be objectively evaluated by experimental methodologies. Results need to be reproducible and comparable. Such widely accepted  methodologies are still crucial challenges.
\end{description}

\section{Analysis}\label{sec:analysis}
% analyse the Siamese network approaches
% show illustrations of specific details (eventually)
Several research groups have been studying very recently Siamese networks for tracking and the literature of this field is rapidly growing. For example, Tao et al.~\cite{tao-CVPR2016} received since 2016 130 citations\footnote{Source: Google Scholar, 12/6/2018}, in 2016: 8, in 2017: 76, and since 2018 until June: 46. Following our argumentation for a better understanding of Siamese networks (Section~\ref{sec:introduction}), this study selects nine papers for a deeper analysis~\cite{fan-TNN2010, tao-CVPR2016, held-ECCV2016, bertinetto-VOT2016, valmadre-CVPR2017, wang-CORR2017, choi-CORR2017, guo-ICCV2017, chen-TCSVT2017}. These papers were published in major journals and conferences between 2016 and 2017, except Fan et al.~\cite{fan-TNN2010} which is to our knowledge the earliest work in this field.

%%%%%%%%%%%%%%%%%%%%%%%%%%%%%%%%%%%%%%%%%%%%%%%%%%%
\subsection{Methodology}
% analyse methods based on the general tracking problem formulation
% tracker function, pre-processing, network, post-processing function (inference)
To understand the results of this analysis, this work needs to explain the methodology for the analysis and to describe some aspects concerning the formalism. The study acts on the problem formulation given by Section~\ref{sec:problem} and identifies the tracker function and its properties (Equation~\ref{eq:tau-n}) with a deep analysis of the trackers chosen in this study. The assumed underlying design of all the trackers follows our suggestion of Equation~\ref{eq:sigma}. Both partial tracker functions as well as the Siamese network function are explained in the following in detail.

% initialisation, input, output of network function
Concerning the network function, we introduce $T_I$ for image $X_{t-1}$ of the sequence of images $X(t)$. Then, we introduce $S_I$ describing $X_t$ and two further names according to $S_I$, namely the search region $S_R$ and the target patch $T_P$ which is matched to the search region. Those patches are derived by a bounding box $B$ which is in all trackers part of $Y(t)$. All trackers assume an initial given bounding box.
 
Target patch and search region are input to the network function, except for Tao et al.~\cite{tao-CVPR2016} who use the whole image and additional proposals of bounding boxes $B_S$\footnote{Based on the idea of region CNNs~\cite{girshick-ICCV2015}} and for Fan et al.~\cite{fan-TNN2010} and Wang et al.~\cite{wang-CORR2017} who use a search patch of the same size as the target patch by assuming small object motion. The proposed network functions output either a probability map $P_M$ or a score map $S_M$ of object location. The main difference is that scores in $S_M$ are unbounded.

\begin{figure*}
\centering
\includegraphics[width=.7\textwidth]{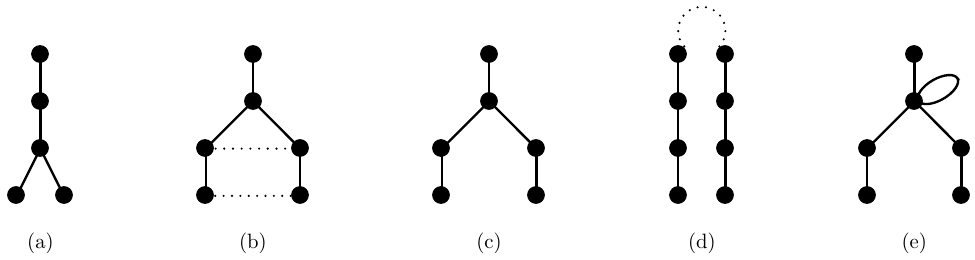}
\caption{The trackers use network architectures of varying complexity: (a) Two-Channel Siamese, (b) Pseudo Siamese, the dotted line indicate shared weights, (c) Siamese, (d) Two-Stream Siamese, the dotted line indicates the shared output space and (e) Recurrent Siamese.}
\label{fig:net}
\end{figure*}

% network function, integral parts - (branches, connection of branches)
The network function matches basically the target patch with the search patch or search region. Following a recent classification~\cite{zagoruyko-CVIU2017}, this study divides Siamese networks into (i) Two-Channel Siamese, (ii) Pseudo Siamese, (iii) Siamese (iv) Two-Stream Siamese and (v) Recurrent Siamese network architectures. While Two-Channel Siamese networks share whole layers in their two branches and therefore can be seen as a single holistic network with two input layers, Pseudo Siamese networks share only the network parameters in their branches, i.e. the input layers are mostly independent. Classical Siamese networks have two independent branches. Two-Stream Siamese networks go here one step further as they omit connecting layers and use instead a normalisation layer in the branches which transforms the output in a common manifold of some measurable space where matching is then formulated. Finally, Recurrent Siamese networks have connections among hidden layers either in the branches or in the connecting layers. Fig.~\ref{fig:net} illustrates this classification.

%%%%%%%%%%%%%%%%%%%%%%%%%%%%%%%%%%%%%%%%%%%%%%%%%%%
% show illustration of the classes of Siamese networks

\begin{figure*}
\centering
\subfloat[YCNN]{\includegraphics[width=.9\textwidth]{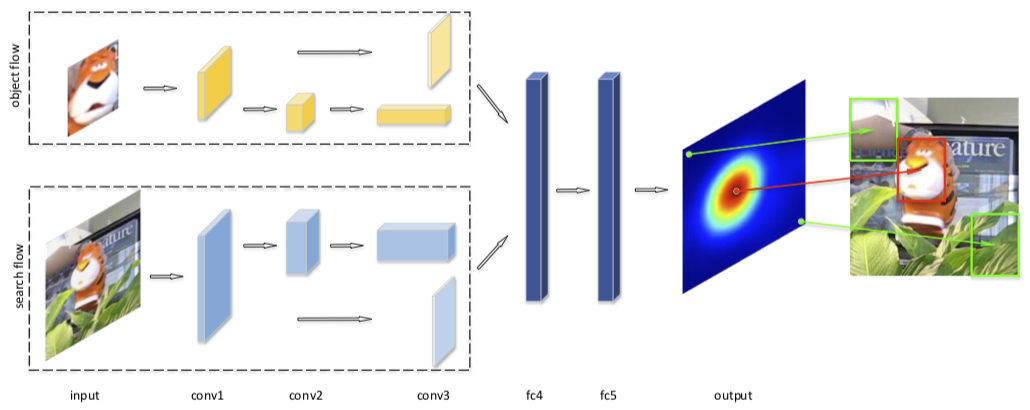}} \\
\subfloat[GOTURN]{\includegraphics[width=.9\columnwidth]{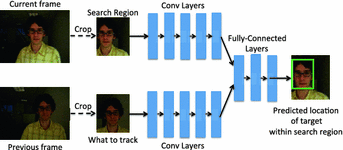}} \qquad
\subfloat[CFNet]{\includegraphics[width=.9\columnwidth]{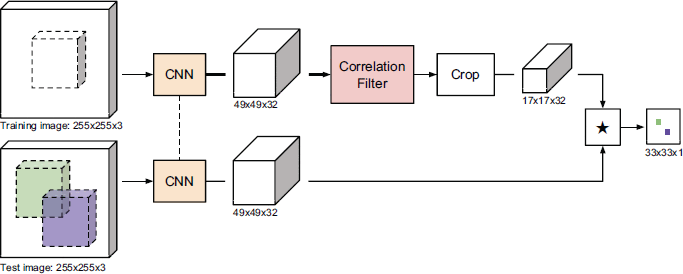}} \\
\caption{An example of a comparison of three trackers by graphical illustrations. Images are from the original papers~\cite{chen-CORR2016, held-ECCV2016, valmadre-CVPR2017}.}
\label{fig:networks}
\end{figure*}

% prefix notation
This analysis identifies the important layers and features for each proposed network function and enumerates them for a better comparison. This is usually done by a graphical illustration of the network (Fig.~\ref{fig:networks}). However, such formalism has its limits as we found it very difficult to understand the differences in the details of the network function, especially when comparing more than two approaches. Another problem is the integration of the network function into the tracker function which is for all papers neglected in the graphical illustrations. As a result, the whole tracker is discussed in an ambiguous way in the text. A contribution of this study is to rewrite the network and the tracker functions by introducing a new formalism based on prefix notation similar to Lisp which allows to compare tracking much easier in a holistic way.

% training
The analysis comprises also the sources of training data as well as the use of training loss and reward function respectively. Training is essential for performance, hence a better comparison and understanding of training loss and data is needed.

%%%%%%%%%%%%%%%%%%%%%%%%%%%%%%%%%%%%%%%%%%%%%%%%%%%
\subsection{Qualitative Analysis}\label{sec:qualitative}
% describe the proposed methods in detail: network (branches, output, connection), training, tracker inference
This section groups the considered trackers by the classification given in Fig.~\ref{fig:net} and analyses qualitatively for each tracker the tracker function (Table~\ref{tab:trackers}) and the network function with its specific training (Table~\ref{tab:networks}). Individual features of the network function as well as the tracker function are identified (Table~\ref{tab:features}). Table~\ref{tab:tracker-details} summarises the specific details of $\delta_0(\cdot)$ and $\delta_1(\cdot)$. Section~\ref{sec:quantitative} shows then progress of research based on a quantitative analysis based on results of VOT and OTB provided by the literature.

% include the results tables
%%%%%%%%%%%%%%%%%%%%%%%%%%%%%%%%%%%%%%%%%%%%%%%%%%%
% table summaries of the qualitative analysis
% summary of network functions
\begin{table}
\caption{Components in State of the Art Siamese Networks and Trackers}
\begin{center}
\begin{tabular}{c|l|p{3cm}}
\hline
Function & Description & Used By \\
\hline
$f_\text{fc}$ & full connection & CNNT, GOTURN, YCNN, RDT \\
$f_\text{c}$ &  convolution/max-pooling & all \\
$f_\text{p}$ & ROI-pooling & SINT \\
$f_\text{w}$ & ridge regression & CFNet, DCFNet \\
$f_\star$ & convolution & DCFNet \\ % $(f_\text{corr}~[P_1~P_2]~(\mathcal{F}^{-1}~(\odot~(\mathcal{F}~P_1)~(\mathcal{F}~P_2))))$
$\star$ & full convolution & SiamFC, CFNet, DSiam \\
$\odot$ & Hadamard & DSiam \\
$l_2$ & Euclidean distance & SINT \\
$f_{l_2}$ & $l_2$ normalisation & SINT \\
$f_\text{avg}$ & mean & CFNet, YCNN, RDT \\
$f_\text{crop}$ & bounding box crop & all except SINT \\
$f_\text{box}$ & bounding box estimator & all \\
$f_{C_i}$ & box corner crop & CNNT \\
$f_\circ$ & isotropic box adaptation & SINT \\
$f_\times$ & box scaling & SiamFC, CFNet, YCNNN, RDT, DSiam \\
$f_\text{ref}$ & box refinement & SINT \\
$f_{\text{scale}_i}$ & patch scaling & SINT, SiamFC, CFNet, YCNN, RDT  \\
$f_{\text{scale}_\sigma}$ & input scaling & DSiam \\
$f_{\text{trans}_i}$ & patch translation & RDT \\
$f_\text{bg}$ & background suppression & DSiam \\
\hline
\end{tabular}
\end{center}
\label{tab:features}
\end{table}

% summary of state-of-the-art networks and training
\begin{table*}
\caption{State of the Art Siamese Networks and Training for Learning Tracking}
\centering
\begin{tabular}{r|p{0.5cm}|p{0.5cm}|p{0.6cm}|p{5cm}|l|p{1.8cm}}
\hline
Method & In & Out & Class & Network Function $\sigma(\cdot)$ & Training Loss/Reward & Training Data \\
\hline
CNNT~\cite{fan-TNN2010} & $T_P$, $S_P$ & $P_M$ & 1 &
$(f_\text{fc}~(f_\text{c}~(f_\text{c}~\theta~T_P~S_P))~(f_\text{c}~(f_\text{c}~\theta~T_P~S_P)))$\textsuperscript{*} &
n/a\textsuperscript{**} &
NEC \\

SINT~\cite{tao-CVPR2016} & $T_I$, $S_I$, $B_T$, $B_S$ & $D_2$ & 4\textsuperscript{***} &
$(l_2~((f_{l_2}~\theta~(f_p~(f_c~T_I)~B_T))~ \newline (f_{l_2}~\theta~(f_p~(f_c~S_I)~B_S)))$ &
$\mathbbm{1}_{B_{T=S}} \sigma(\cdot)^2+\mathbbm{1}_{B_{T\not=S}}\max(0, \epsilon-\sigma(\cdot)^2)$ &
ImageNet-12, ALOV \\

GOTURN~\cite{held-ECCV2016} & $T_P$, $S_R$ & $B$ & 3 & $(f_\text{fc}~(f_\text{c}~T_P)~(f_\text{c}~S_R))$ &
$l_1(\sigma(\cdot), B)$ &
ImageNet-14, ALOV\\

SiamFC~\cite{bertinetto-VOT2016} & $T_P$, $S_R$ & $S_M$ & 3 & $(\star~(f_\text{c}~T_P)~(f_\text{c}~S_R))$ &
$\frac{1}{|S_M|}\sum \log(1+\exp(-\sigma(\cdot) \odot S_M))$ &
ImageNet-15 Video\\

CFNet~\cite{valmadre-CVPR2017} & $T_P$, $S_R$ & $S_M$ & 2 & $(\star~(f_\text{w}~(f_\text{c}~\theta~T_P))~(f_\text{c}~\theta~S_R))$ &
%\newline $\argmin{W} ~l_2(W \star f_c(T_P, \theta),S_M)^2+\lambda \|\vecop{W}\|^2$ &
as SiamFC &
ImageNet-15 Video \\

YCNN~\cite{chen-CORR2016} & $T_P$, $S_R$ & $P_M$ & 2 & $(f_\text{fc}~(f_\text{c}~\theta~T_P)~(f_\text{c}~\theta~S_R))$ &
$l_2(\sigma(\cdot), P_M)^2$ &
ImageNet-12, ALOV, VOT-15\textsuperscript{}, TB-100\textsuperscript{} \\

RDT~\cite{choi-CORR2017} & $T_P$, $S_R$ & $P_M$, $P$ & 2 & $(f_\text{fc}~(f_\text{c}~(\sigma_\text{YCNN}~T_P~S_R)))$ &
$\mathbbm{1}_{\text{success}}(\sigma(\cdot)) - \mathbbm{1}_{\text{failure}}(\sigma(\cdot))$\textsuperscript{****} &
ImageNet-15, VOT-15 \\

DCFNet~\cite{wang-CORR2017} & $T_P$, $S_P$, $W$ & $S_M$, $W$ & 5\textsuperscript{***} &
$(f_\star~(f_\text{w}~(W~(f_\text{c}~T_P)~(f_\text{c}~T_P)))~(f_\text{c}~S_P))$ &
$l_2(\sigma(\cdot, \theta), S_M)^2+\lambda \|\theta\|^2$ &
NUS-PRO, TempleColor128, UAV123 \\

DSiam~\cite{guo-ICCV2017} & $T_P$, $S_R$, $U$, $V$ & $S_M$ & 3 & $(\odot~Y~(\star~(*~U~(f_\text{c}~T_P)~(*~V~(f_\text{c}~S_R)))))$ &
as SiamFC &
ImageNet-15 Video \\
\hline
\end{tabular}
\\[1ex]
\parbox{\textwidth}{\itshape \footnotesize \textsuperscript{*}~Parameters $\theta$ are shared across feature transforms \quad \textsuperscript{**} "Difference" between probability maps\cite{fan-TNN2010} \quad \textsuperscript{***} Weight sharing introduces properties of Pseudo Siamese networks  \quad \textsuperscript{****} Episode reward \quad In: $T_I$ = Target Image, $S_I$ = Search Image, $T_P$ = Target Patch,  $S_P$ = Search Patch, $S_R$ = Search Region, $U$ = Appearance Transform, $V$ = Background Suppression \quad Out: $P_M$ = Probability Map, $D_2$ = $l_2$ Distance, $B$ = Bounding Box, $S_M$ = Score Map, $P$ = Probability \quad In/Out: W = Correlation Kernel \quad Class: 1 = Two-Channel Siamese, 2 = Pseudo Siamese, 3 = Siamese, 4 = Two-Stream Siamese, 5 = Recurrent Siamese}
\label{tab:networks}
\end{table*}

% summary of tracker functions
\begin{table}
\caption{State of the art visual tracking with Siamese networks}
\begin{center}
\begin{tabular}{r|c|p{.5cm}|l}
\hline
Method & Class & $Y(t)$ & Tracker Function $\tau(t)$ \\
\hline
CNNT & 2M\textsuperscript{*} &
$B$ &
$(\delta_1~(\delta_0~T_I^t~T_I^{t-1}~B^{t-1}))$ \\

SINT & 1 &
$B$ &
$(\delta_1~(\delta_0~T_I^t~T_I^0~B^{t-1}~B^0))$ \\

GOTURN & 2M &
$B$ &
as CNNT \\

SiamFC & 1 &
$B$ &
as SINT \\

CFNet  & 2 &
$B$ &
$(\delta_1~(\delta_0~\{ T_I^{i-1}, T_I^i \}_{i\le t}~\{ B^{i-1}, B^i \}_{i<t}))$ \\

YCNN & 2 &
$B$ &
$(\delta_1~(\delta_0~\{T_I^i\}_{i\le t}~\{B^i\}_{i<t}))$ \\

RDT & 2 &
$B$ &
as YCNN \\

DCFNet & 2M &
$B$, $W$ &
as CNNT \\

DSiam & 1 &
$B$, $U$, $V$ &
as SINT \\
\hline
%\multicolumn{4}{c}{\rule{0pt}{2ex} \footnotesize{\textsuperscript{*}~$\forall i<t, \exists j: i+1 = j$}} \\
\end{tabular}
\\[1ex]
\parbox{\textwidth}{\itshape \footnotesize \textsuperscript{*} 2-Type tracker with Markov assumption}
\end{center}
\label{tab:trackers}
\end{table}

% summary of the trackers' details
\begin{table*}
\caption{State of the art visual tracking with Siamese networks (cont.)}
\begin{center}
\begin{tabular}{l|p{8cm}|p{7cm}}
\hline
Method & Partial Tracker Function $\delta_0(\cdot)$ & Partial Tracker Function $\delta_1(\cdot)$ \\
\hline

CNNT &
$(\text{setv}~T_P~(f_\text{crop}~T_I^{t-1}~B^{t-1})~S_P~(f_\text{crop}~T_I^t~B^{t-1}))$ \newline
$(\text{setv}~T_P^i~(f_\text{crop}~T_I^{t-1}~(f_{C_i}~B^{t-1}))~S_P^i~(f_\text{crop}~T_I^t~(f_{C_i}~B^{t-1})))$ &
$(f_\text{box}~(f~(\sigma_\text{CNNT}~\theta~\cdot)~\{ (\sigma_\text{CNNT}~T_P^i~S_P^i~\theta) \}_{1\le i\le 4}))$\textsuperscript{*} \\

SINT &
$(\text{setv}~B_T ~B^0~B_S~(f_\circ~B^{t-1}))$ &
$(f_\text{ref}~(f_\text{box}~(\argmin{} ~\{ (f_{\text{scale}_i}~B_S) \}_i ~(\sigma_\text{SINT}~\cdot))))$ \\

GOTURN &
$(\text{setv}~T_P~(f_\text{crop}~T_I^{t-1}~B^{t-1})~S_R~(f_\text{crop}~T_I^t~B^{t-1}))$ &
$(\sigma_\text{GOTURN}~\cdot)$ \\

SiamFC &
$(\text{setv}~T_P~(f_\text{crop}~T_I^0~B^0)~S_R~(f_\text{crop}~T_I^t~(f_\times~5~B^{t-1})))$ &
$(f_\text{box}~(\max~\{ (\sigma_\text{SiamFC}~(f_{\text{scale}_i}~T_P)~\cdot) \}_i))$ \\

CFNet &
$(\text{setv}~T_P~(f_\text{avg}~\{ (f_\text{crop}~T_I^i~B^i) \}_{i<t})~S_R~(f_\text{crop}~T_I^t~(f_\times~4~B^{t-1})))$ &
as SiamFC \\

YCNN &
$(\text{setv}~T_P~\{ (f_\text{crop}~T_I^i~B^i) \}_{i<t}~S_R~(f_\text{crop}~T_I^t~(f_\times~2.5~B^{t-1})))$ &
$(f_\text{box}~(\max~(f_\text{avg}~\{ (*~\mu_{ij}~(\sigma_\text{YCNN}~(f_{\text{scale}_i}~T_P^j)~\cdot) \}_{i, j}))))$ \\

RDT &
as YCNN &
$(f_\text{box}~\{ (\sigma_\text{YCNN}~(f_{\text{trans}_i}~(f_{\text{scale}_j}~T_P^k)~\cdot) \}))$ s.t. \newline
$(\argmax{}~\{ (f_{\text{trans}_i}~(f_{\text{scale}_j}~T_P^k) \}_{1\le i, j\le 3\le i\le 4, k})~(\sigma_\text{RDT}~\cdot))$ \\

DCFNet &
$(\text{setv}~T_P~(f_\text{crop}~T_I^{t-1}~B^{t-1})~S_P~(f_\text{crop}~T_I^t~B^{t-1}))$ &
$(f_\text{box}~(\text{get}~(\sigma_\text{DCFNet}~\cdot)~B))$\textsuperscript{**} \newline
$(\text{get}~(\sigma_\text{DCFNet}~\cdot)~W)$ \\

DSiam &
$(\text{setv}~T_P~(f_\text{crop}~T_I^0~B^0)~S_R^i~\{ (f_\text{crop}~T_I^t~(f_\times~\gamma_i~B^{t-1})) \}_i)$\textsuperscript{***} \newline
$(\text{setv}~U~(f_\text{w}~U^{t-1}~(\sigma_\text{DSiam}~T_P~\theta~\cdot)~(\sigma_\text{DSiam}~(f_\text{crop}~T_I^{t-1}~B^{t-1})~\cdot)))$ &
$(f_\text{box}~(\max~\{ (\sigma_\text{DSiam}~(f_{\text{scale}_\sigma}~S_R^i)~\cdot) \}_{1\le i \le 3}))$ \newline
$(\text{setv}~V~(f_\text{w}~V^{t-1}~(\sigma_\text{DSiam}~S_R^2~\theta~\cdot)~(\sigma_\text{DSiam}~(f_\text{bg}~S_R^2~B)~\cdot)))$ \\
%\argmin{U}~l_2(U\star \sigma_\text{DSiam}(T_P^0, \theta), \sigma_\text{DSiam}(T_P, \cdot))^2+\lambda \| \vecop{U} \| ^2
%\argmin{V}~l_2(V\star \sigma_\text{DSiam}(S_R^2, \theta), \sigma_\text{DSiam}(f_\text{bg}(S_R^2, B), \cdot))^2+\lambda \| \vecop{V} \| ^2
\hline
\multicolumn{3}{c}{
\begingroup
\renewcommand{\arraystretch}{}
\begin{tabular}{c}
\footnotesize{\textsuperscript{*}~$f(\cdot)$ n/a \quad \textsuperscript{**}~application of $f_\text{scale}$ unspecified. "We use patch pyramid..." \cite{wang-CORR2017}. \quad \textsuperscript{***}~$\gamma_2 = 1$} \\
\end{tabular}
\endgroup}
\end{tabular}
\end{center}
\label{tab:tracker-details}
\end{table*}

%%%%%%%%%%%%%%%%%%%%%%%%%%%%%%%%%%%%%%%%%%%%%%%%%%%
\subsubsection{Two-Channel Siamese Networks (CNNT)}
This work~\cite{fan-TNN2010}  is inspired by work on convolutional neural networks from the 90s~\cite{matan-NIPS1991, nowlan-NIPS1994}. $\sigma_\text{CNNT}(\cdot)$ is a Two-Channel Siamese network and proposes two input branches $f_c$, i.e. a single convolutional (conv-1) and max-pooling (pool-1) layer, followed by conv-2, pool-2, conv-3, all sharing the weights $\theta$, and a final four-times upsampling layer. This four-times upsampling makes the network translation variant which is to emphasise. Conv-2 works individually for each branch, but connects also feature maps of both branches to a joint third branch by a pre-defined convolutional scheme. Connections between conv-2 and pool-2 are random. A fourth branch connects pool-1 directly with conv-4 followed by a translation transform layer. The output layer $f_\text{fc}$ adds these four branches followed by a sigmoid to create a probability map $P_M$ of location. The input are two normalised image patches $T_P$ and $S_P$ (RGB and two channels of the image gradient), one centred at the target position in the previous image and one showing in the same position the new image information in the current image of the sequence.
$\sigma_\text{CNNT}(\cdot)$ captures multilayer features by (i) the two input branches exploiting spatial fine-to-coarse details of $T_P$ and $S_P$, (ii) the third internal branch keeping the fine local spatial details and (iii) the forth branch capturing fine-to-coarse spatiotemporal information, therefore the network discriminates jointly the appearance and motion of objects.
Training is shown for persons' heads in varying poses and difficulties (changes in illumination and view) on a proprietary person dataset provided by the authors (20k samples). The parameters of the network are trained offline by minimising a difference between ground truth probability maps and network output and by using stochastic gradient descent. Unfortunately, details about the loss function are not given in the paper. 
Tracking implements a feed forward through the network of the normalised online data starting with the initial bounding box of a person's head. The new position is then simply set to the largest peak in the probability map. Hence, CNNT is of Type-2M as it immediately forgets image patches of previous times. To face scale change, CNNT uses two further CNNs similar to $\sigma_\text{CNNT}(\cdot)$ to track the four keypoints ($f_{C_i}$) of the bounding box.

%%%%%%%%%%%%%%%%%%%%%%%%%%%%%%%%%%%%%%%%%%%%%%%%%%%
% show trackers using a target template and search patch
%\begin{figure}
%\centering
%\subfloat[CNNT]{\includegraphics[width=.9\columnwidth]{5559504-fig-3-large}} \\
%\subfloat[DCFNet]{\includegraphics[width=.9\columnwidth]{dcfnet}}
%\caption{Target template and search patch as input.}
%\label{fig:TTSP}
%\end{figure}

%%%%%%%%%%%%%%%%%%%%%%%%%%%%%%%%%%%%%%%%%%%%%%%%%%%
\subsubsection{Pseudo Siamese Networks}
\begin{description}
\item[CFNet] builds upon CFNet and adds a correlation filter layer $f_w$ to the branch which processes $T_P$~\cite{valmadre-CVPR2017}. These layers follow directly the convolutional network $f_c$. $\sigma_\text{CFNet}(\cdot)$ shares weights between the two branches (Pseudo Siamese) which is not the case for SiamFC. The input to the other branch is a larger region of the image $S_R$ including the object, hence resolution of feature maps in the branches and score map is larger. Feature maps are further multiplied by a cosine window and cropped after correlation to remove the effect of circular boundaries by $f_w$.
CFNet inherits the basic ability from SiamFC to discriminate spatial features with given triplets of target patch, search region and corresponding score map. Instead of unconstrained features, CFNet learns features that especially discriminate and solve the underlying ridge regression $f_w$ by exploiting background samples in the surrounding region of the object. The weights of the correlation layer remain constant during tracking, no online learning happens as shown by Danelijan et al.~\cite{danelljan-ICCVW2015}.
Training is done as with SiamFC by using the same loss function and optimisers on videos of objects from ImageNet~\cite{russakovsky-IJCV2015}. An emphasis of the work is put to the back-propagation and the differentiability of $f_w$ for the end-to-end training. The formulation in Fourier space preserves efficiency of computation.
Tracking is as simple as in SiamFC. A forward pass through $\sigma_\text{CFNet}(\cdot)$ computes position and scale of the object. The score map is multiplied by a spatial cosine window to penalise larger displacements. Instead of handling five different scale variations, scale $f_{\text{scale}_i}$ is handled similar to Comaniciu et al.~\cite{comaniciu-TPAMI2002}. To fully exploit the correlation filter, the initial template is updated in each frame by a moving average, therefore CFNet is against SiamFC a Type-2 tracker.

\item[YCNN] proposes a Pseudo Siamese network with two branches similar to VGGNet~\cite{simonyan-ICLR2015} with three convolutional and max-pooling layers $f_c$, both branches connected to three fully-convolutional layers $f_\text{fc}$~\cite{chen-CORR2016}. The convolutional layers share the same weights $\theta$. Each layer finishes with a ReLU except for the output which finishes with a sigmoid function. The network output is a probability map $P_M$ with values near to one at pixels indicating the object's presence.
Again, the branches work as feature hierarchies aggregating fine-to-coarse visual details $f_\text{fc}$ captures spatiotemporal information and describes similarity. YCNN learns discriminating features of growing complexity while simultaneously learning similarity between target patch and search region with corresponding probability maps. We emphasise here that no further assumptions on similarity are given. It is solely descriptively given by the training samples.
Training is done in two stages on augmented images of objects from ImageNet~\cite{krizhevsky-NIPS2012} and for fine-tuning with videos from ALOV~\cite{smeulders-TPAMI2013}, VOT-2015~\cite{kristan-ICCVW2015} and TB-100~\cite{wu-TPAMI2015}. Training minimises a weighted $l_2$ loss by using Adam~\cite{Kingma-CORR2014}, mini-batches and dropout. Weighting is important as nearly 95\,\% of pixels in the prediction map have very low to zero values.
YCNN is a Type-2 tracker, as it maintains a pool of target patches. In contrast to CFNet these patches need not to be from subsequent images. $\sigma_\text{YCNN}(\cdot)$ infers position as the largest probability in $P_M$. By averaging $f_\text{avg}$ the probability map over the five most confident target patches avoids drift. Repeating inference with scaled patches by $f_{\text{scale}_i}$ estimates additionally overall scale.

\item[RDT] proposes as network $\sigma_\text{RDT}(\cdot)$ which combines $\sigma_\text{YCNN}(\cdot)$ with a convolutional policy network to implement temporal difference learning~\cite{choi-CORR2017}. In principle any kind of matching network could be used. The policy network $f_c$ has two convolutional layers with ReLUs and a final max pooling layer, followed by two fully connected layers with dropout normalisation and a sigmoid. The inputs are the target patch $T_P$ and a search region $S_R$. The output of $\sigma_\text{YCNN}(\cdot)$ is a probability map $P_M$ which is input to $\sigma_\text{RDT}(\cdot)$. Final output is a probability of the policy currently used, i.e. the chosen target patch.
The idea is to decide for a target patch out of a pool of target patches which gives largest probability. This decision is done for each inference done with $\sigma_\text{YCNN}(\cdot)$. The pool is adapted over time, however, patches such as the initial target patch can remain in the pool as long as they are important for the policy. $\sigma_\text{RDT}(\cdot)$ estimates basically reliability of the target patches and forces the pool to be diverse as possible~\cite{nebehay-MSC2013, meshgi-CVPR2018}.
A policy is learned based on rewards given after whole episodes of videos from ImageNet~\cite{russakovsky-IJCV2015} and VOT-2015~\cite{kristan-ICCVW2015}. Instead of Q-statistics, the policy function is here differentiable. To improve convergence, a number of successful and erroneous episodes is kept to update the gradient after each episode. Failure is determined by IoU under 0.2 averaged over the last 30 bounding boxes of an episode. The matching network $\sigma_\text{YCNN}(\cdot)$ is trained by using Adam~\cite{Kingma-CORR2014}.
RDT is like YCNN a Type-2 tracker. Tracking estimates for all patches in the pool (max. four patches) their policy score. The patch with the largest policy score has also been fed into $\sigma_\text{YCNN}(\cdot)$. Position of the object is then given by $P_M$ at the pixel with largest probability. To account for scale, $f_{\text{scale}_i}$ gives three target patches which are considered. RDT uses also translated patches by $f_{\text{trans}_i}$. A patch in the pool is randomly replaced by the new patches every 50 frames.
\end{description}

%%%%%%%%%%%%%%%%%%%%%%%%%%%%%%%%%%%%%%%%%%%%%%%%%%%
\subsubsection{Siamese Networks}
\begin{description}
\item[GOTURN] proposes a classical Siamese network $\sigma_\text{GOTURN}(\cdot)$ with two convolutional branches $f_c$ inherited from AlexNet up to pool-5~\cite{held-ECCV2016}. The resulting pool-5 feature maps of both branches are fed to three fully-convolutional layers $f_\text{fc}$ which use ReLUs after each layer. The final output layer yields a vector describing directly the bounding box $B$. The output is scaled by a validated constant factor. 
GOTURN learns simultaneously the hierarchy of spatial features in the branches as well as spatiotemporal features and the similarity function in the fc layers to discriminate between target patch $T_P$ and search region $S_R$ with corresponding bounding boxes.
Training is done in two stages on augmented images of objects from ImageNet~\cite{russakovsky-CORR2014} and on videos from ALOV~\cite{smeulders-TPAMI2013} by using standard back-propagation of CaffeNet~\cite{jia-CORR2014}. Augmentation assumes linear translation and constant scale with parameters sampled from a Laplace distribution, hence small motion is assumed to occurs more frequently than larger displacements. Training minimises a $l_1$ loss between predicted and ground truth bounding box by using mini-batches, dropout and pre-training of the branches on ImageNet without fine-tuning these parameters to prevent overfitting. 
GOTURN is as CNNT a classical Type-2M tracker as it initialises $T_P$ with the first image and updates the target patch with the predicted bounding box for each image in the sequence. Crops of the current and next image in the sequence yield $T_P$ and $S_R$. These crops are not exact but padded to add context information of the background.

\item[SiamFC] proposes a Siamese network $\sigma_\text{SiamFC}(\cdot)$ with two identical branches $f_c$ inherited from AlexNet with five convolutional layers, then max-pooling after conv-1 and conv-2 and ReLUs after every convolutional layer except for conv-5~\cite{bertinetto-VOT2016}. A novel cross-correlation layer $f_\star$ connects conv-5 of the two branches. By waiving padding the whole network is fully-convolutional, meaning that beside the target patch $T_P$, a search region $S_R$ is allowed to be matched instead of a search patch such as for CNNT which is to emphasise. The output is an unbounded correlation or score map $S_M$ with high values at pixel locations indicating high likelihood of object presence.
Similar to SINT, the branches can be seen as transformation of visual descriptions with increasing spatial receptive fields whereas the final feature maps are embedded in a measurable space. Here, cross-correlation is used as similarity function. SiamFC learns discriminating solely the features with given triplets of target patch, search region and corresponding score map. Values isotropically within a radius of the centre count correctly to the object's position, hence are labeled positively whereas all other values are labeled negatively.
Training is done on videos of objects from ImageNet~\cite{russakovsky-IJCV2015}. Augmentation considers scale but not translation, because of the fully-convolutional network property. Training minimises a discriminative mean logistic loss by using SGD, mini-batches, Xavier initialisation and annealing of the learning rate. 
Tracking is of Type-1 as it computes the position via the up-sampled score map for a given initial patch. The tracker handles scale by searching over five different scale variations and updates scale by linear interpolation $f_{\text{scale}_i}$.

\item[DSiam] introduces a Siamese network $\sigma_\text{DSiam}(\cdot)$ similar to SiamFC which adopts the branches either from AlexNet or VGGNet up to pool-5~\cite{guo-ICCV2017}. The branches finish by layers of regularised linear regression and a preceding element-wise multi-layer fusion of conv-4 and conv-5 which is inspired by Ma et al.~\cite{ma-ICCV2015}. The branch concerning the target patch regresses an affine transformation $U$ between initial and current patch, whereas the branch concerning the search region adapts a further affine transformation $V$ that suppresses all features of the search region similar to the features of the current target patch ($f_\text{bg}$). Initial target patch $T_P$ and search region $S_R$ containing the object in the current image are input to $\sigma_\text{DSiam}(\cdot)$. The output after correlation is a score map $S_M$ where peaks indicate the center of the object. In case of multi-layer fusion, the final score map is an element-wise weighted sum of the correlations at different depths of the feature hierarchy.
DSiam learns deep features in the respective branches, however, it has in contrast to SiamFC the additional capability to online adapt affine changes of the initial patch and at the same time to suppress object features in the search region which are in the background which is here clearly to emphasise. Fusion of conv-4 and conv-5 allows equal contribution of features showing fine and coarse details.
Training is done for all components jointly in an end-to-end manner on 2000 video clips containing 10 frames each, generated from ImageNet~\cite{russakovsky-IJCV2015}. The momentum optimiser uses a logistic loss function terminated after 50 iterations.
DSiam is a conservative Type-1 tracker, i.e. the initial target patch is always exploited. Instead of scaling $T_P$ by $f_{\text{scale}_i}$, DSiam computes three response maps according to three different scales of $S_R$ by $f_{\text{scale}_\sigma}$ which is to emphasise. $U$ is directly computed by using the initial patch and the target patch of the previous image. The largest score in the score map yields then the new bounding box which is input to the computation of $V$.
\end{description}

%%%%%%%%%%%%%%%%%%%%%%%%%%%%%%%%%%%%%%%%%%%%%%%%%%%
\subsubsection{Two-Stream Siamese Networks (SINT)}\label{sec:sint}
Tao et al.~\cite{tao-CVPR2016} propose two identical query and search networks inherited from AlexNet~\cite{krizhevsky-NIPS2012} and VGGNet~\cite{simonyan-ICLR2015} with five convolutional and two max-pooling layers ($f_c$), three region pooling layers $f_p$, and a final normalisation layer $f_{l_2}$. ReLUs follow each convolution. Max-pooling is done after conv-1 and conv-2. Both networks share the same weights $\theta$ which classifies the network somewhere between Pseudo Siamese and Two-Stream. Two subsequent images, the query image $T_I$ and the search image $S_I$ are input, therefore additional bounding boxes $B_T$ locating the object in $T_I$ and bounding boxes $B_S$ locating candidate objects in $S_I$ are fed to $\sigma_\text{SINT}(\cdot)$. The network yields two normalised features lying on the same manifold in a measurable space such as the $l_2$-space.
$\sigma_\text{SINT}(\cdot)$ aggregates fine-to-coarse spatial details of the object in a feature hierarchy, but without learning jointly a similarity, as similarity is basically given by the training loss which is to emphasise.
Training is done on images of objects from ALOV\cite{smeulders-TPAMI2013}. Training minimises a margin contrastive loss and uses pre-training on ImageNet~\cite{krizhevsky-NIPS2012}. The margin contrastive loss discriminates features of growing complexity with bounding boxes in query and search frame and an additional binary variable indicating correct and incorrect pairs measured by the Jaccard index.
The network is always fed with the initial bounding box and the first image of the sequence which results in a query feature vector. Therefore, SINT is classified as Type-1. $f_\circ$ samples candidates at radial positions and $f_{\text{scale}_i}$ scales differently to feed $B_S$ all at once to the network, resulting in feature vectors for each candidate bounding box. An offline learned ridge regressor $f_\text{ref}$ refines finally position and scale of the winning candidate with maximal inner product to the query. $f_{\text{scale}_i}$ is improved by a variant of SINT called SINT+. Here, the sampling range is adaptive~\cite{wang-ICCV2015} and optical flow is used to filter out false candidates.

%%%%%%%%%%%%%%%%%%%%%%%%%%%%%%%%%%%%%%%%%%%%%%%%%%%
\subsubsection{Recurrent Siamese Networks (DCFnet)}
Wang et al.~\cite{wang-CORR2017} proposes in the network $\sigma_\text{DCFNet}(\cdot)$ a correlation layer $f_\star$ as connecting layer. The two branches of the network are lightweight by stopping for both at conv-1 of AlexNet without max-pooling and with a reduced number of 32 feature maps before the correlation. Inputs are the centred target patch $T_P$ and the search patch $S_P$, both equal in size. This is in contrast to e.g. SiamFC which uses a fully convolutional layer to allow search regions. The correlation filter is integrated in the target patch branch and is updated online by feeding the kernel of the target patch at the previous time step into $f_w$. This makes $\sigma_\text{DCFNet}(\cdot)$ a recurrent network. The output of the network is a score map.
DCFnet is a first attempt to use a recurrent architecture. The kernel acts like a memory and is adapted over time by exponential averaging. The feedback connections provide nominator and denominator of the closed-form regression in $f_w$ of the previous time step.
Training is done by using a regularised $l_2$ loss and stochastic gradient decent on patch pairs coming from NUS-PRO~\cite{li-TPAMI2016}, TempleColor128~\cite{liang-TIP2015} and UAV123~\cite{mueller-ECCV2016}.
DCFNet is implemented as Type-2M tracker, i.e. starting from the initial target patch the tracker adapts after the inference in each step the target patch. Scale variation is accounted by using a pyramid of the input, but the paper does not give any further details.

%%%%%%%%%%%%%%%%%%%%%%%%%%%%%%%%%%%%%%%%%%%%%%%%%%%
% show SINT
%\begin{figure}
%\centering
%\includegraphics[width=.9\columnwidth]{7780527-fig-2-large}
%\caption{SINT}
%\label{fig:SINT}
%\end{figure}

%%%%%%%%%%%%%%%%%%%%%%%%%%%%%%%%%%%%%%%%%%%%%%%%%%%
% show trackers using a target patch 
%\begin{figure}
%\centering
%\subfloat[GOTURN]{\includegraphics[width=.9\columnwidth]{419956_1_En_45_Fig2_HTML}} \\
%\subfloat[YCNN]{\includegraphics[width=.9\columnwidth]{ycnn}} \\
%\subfloat[RDT]{\includegraphics[width=.9\columnwidth]{rdt}}
%\caption{Target Patch and Search Region as Input. Connects by $f_\text{fc}$.}
%\label{fig:TTSRFC}
%\end{figure}

%%%%%%%%%%%%%%%%%%%%%%%%%%%%%%%%%%%%%%%%%%%%%%%%%%%
%\begin{figure}
%\centering
%\subfloat[SiamFC]{\includegraphics[width=.9\columnwidth]{siamfc}} \\
%\subfloat[CFNet]{\includegraphics[width=.9\columnwidth]{8100014-fig-1-large}} \\
%\subfloat[DSiam]{\includegraphics[width=.9\columnwidth]{8237458-fig-2-large}} \\
%\caption{Target Patch and Search Region as Input. Connects by $\star$, $f_\star$.}
%\label{fig:TTSRC}
%\end{figure}

%%%%%%%%%%%%%%%%%%%%%%%%%%%%%%%%%%%%%%%%%%%%%%%%%%%
% 2018
% network
%input/output
%interpretation
% training
% tracking

%%%%%%%%%%%%%%%%%%%%%%%%%%%%%%%%%%%%%%%%%%%%%%%%%%%
\subsection{Quantitative Analysis}\label{sec:quantitative}
% acquire comparable measures for VOT and OTB
After the previous in-depth qualitative analysis, this section compares trackers based on quantitative measures. Our study focuses on OTB and VOT data. Most of the work uses OTB and VOT for evaluation. We considered papers from the existing literature for this study including all key papers~\cite{fan-TNN2010, tao-CVPR2016, held-ECCV2016, bertinetto-VOT2016, valmadre-CVPR2017, wang-CORR2017, choi-CORR2017, guo-ICCV2017, chen-TCSVT2017}, dozens of very recent, directly referring papers and existing surveys with empirical evaluation~\cite{smeulders-TPAMI2013, liang-TIP2015, li-PR2018}. The result is summarised by Table~\ref{tab:comparison} which collects finally comparable AuC values and A/R-scores from several papers~\cite{he-CVPR2018, wang-CVPR2018, zhang-NC2018, guo-ICCV2017, li-PR2018}. Some key papers provide quantitative measurements for multiple trackers.

% availability is a problem
Although literature has been quickly growing for the last two years, as mentioned at the beginning of Section~\ref{sec:analysis}, it was impossible to identify the papers and a particular dataset in literature, neither for OTB nor for VOT, which allows to collect comparable measurements for all the trackers in this evaluation. One recognises by going through all the papers that OTB-2013~\cite{wu-CVPR2013} and VOT-2015~\cite{kristan-ICCVW2015} are the most frequently used benchmarks in literature. VOT and OTB were introduced in 2013 and one reason is that data was not available for trackers introduced before 2013.

But there is another severe reason. VOT challenges add each year a new dataset to the consequently increasing number of benchmarks. Research tends to use the newest dataset, we believe because of the attractiveness of the latest challenge with the negative effect that results scatter increasingly over the benchmarks. What is good for a yearly competition, is counterproductive for comparing seriously a growing number of trackers over many years. The situation for OTB is better. Three datasets are available (OTB-2013, TB-50, TB-100) and the evaluation results in literature scatter over mostly OTB-2013 and TB-100 as TB-50 is in literature ignored or sometimes mistaken for TB-100. 

% comparison of results for OTB-2013, TB-100
Considering OTB-2013, TB-100, Table~\ref{tab:comparison} shows for all trackers a margin (5\,\% OTB-2013, 7\,\% TB-100) to the best performing reference trackers CCOT\_CFCF~\cite{kristan-VOT2017, gundogdu-TIP2018} and MDNet~\cite{nam-CVPR2016}. These results do not show evidence towards certain decisions on the design of tracking except for bounding box regression as for GOTURN which shows a clear inferiority in performance against all other trackers. DSiam is best performing (64.2), while RDT gives the largest AuC value for TB-100 (60.3). TB-50 does not provide enough data for comparison.

Type-1 trackers perform theoretically for fair videos on average more robust as Type-2 trackers for single-pass experiments (OTB OPE), because Type-1 trackers always keep the labelled initial description of the object while Type-2 trackers' descriptions loose this information over time due to the adaption. DSiam (64.2) and SINT (62.5) show indeed best performance supporting this hypothesis, nevertheless, this evidence is weak as an explicit robustness measurement is not provided by the AuC values. OTB-2013 and TB-100 AuC values are clearly below 70\,\% for all trackers.

% comparison of results for VOT-2015
Considering VOT-2015, the available measurements in the papers when compared are very uncertain and raw A/R-scores are often unavailable. The ranking scores are useful to order the trackers in a challenge, but ranking scores are not useful to compare trackers over years as the rank is a relative measure for a specific challenge. We recognise by comparing the experimental setup in the papers that R-scores are very sensitive to changes in training and evaluation data. The AuC values instead absorb such changes by integration of the success curve. R-scores are heavily fluctuating between very large values for CFNet (2.52) and moderate values such as for GOTURN (0.2). The R-score for MDNet also varies significantly between the original paper (0.69) and the paper of Guo et al.~\cite{guo-ICCV2017} (0.36) who showed that training with the test dataset yield the scores in the original paper. A-scores are clearly below 60\,\% for all trackers. Verification by the VOT organisers (A/R-score = 0.63/0.16) is only available for VOT-2016 as well as the measurements for the second reference CCOT\_CFCF\footnote{with additional HOG and color-name features.} (A/R-score = 0.54/0.63).

%Efficiency
% CFNet seems a promising method as it has state-of-the-art performance with 75\,fps\footnote{GOTURN runs best at 100\,fps \cite{held-ECCV2016}.} with less than 4\,\% of the parameters of other five layer methods (in total 600\,kB) such as SiamFC. This makes CFNet applicable to embedded applications. CFNet, SiamFC and SINT show comparable performance by reaching IoU/prec. 60/80\,\% on OTB-2013 and one pass evaluation (OPE)\footnote{SINT performs best with IoU/prec. 62.5/84.8\,\% on OTB-2013 and OPE \cite{tao-CVPR2016}.}. \cite{chen-CORR2016} reports significant lower IoU/prec. of 60/70\,\%. \cite{held-ECCV2016} did not report results on OTB-2013.

% tabular summary of the quantitative analysis results
\begin{table}
\caption{Results of the Quantitative Analysis}
\centering
\begin{tabular}{r|r|r|r|r|r}
\hline
 & OTB-13 & TB-50 & TB-100 & \multicolumn{2}{c}{VOT-15} \\
\hline
Tracker & \multicolumn{3}{c|}{AuC} & A-score & R-score \\
\hline
CNNT & n/a & n/a & n/a & n/a & n/a \\
SINT & 62.5 & n/a & $^5$58.3 & n/a & n/a \\
GOTURN & $^1$44.7 & n/a & $^4$41.0 & $^1$0.51 & $^1$0.20  \\ %VOT14 0.39, 0.1
SiamFC & $^3$60.7 & $^3$51.6 & $^2$58.2 & $^1$0.53 & $^1$0.29 \\
CFNet & $^3$61.1 & $^3$53.0 & $^2$56.8 & $^2$0.56 & $^2$ 2.52 \\
YCNN & $^6$60.1 & 47.6 & 52.9 & n/a & n/a \\ % OTB-2013 from arXiv paper
RDT& n/a & 65.4 & 60.3 & n/a & n/a \\
DCFnet & 60.4 & n/a & $^2$58.0 & $^2$0.55 & $^2$1.59 \\
DSiam & 64.2 & n/a & 57.5 & 0.54 & 0.28 \\ %VOT15
\hline
CCOT\_CFCF & 69.2 & n/a & 67.8 & n/a & n/a \\ % VOT2016 (with other features)
MDNet & n/a & 70.8 & 67.8 & 0.6 & 0.69 \\
$^1$MDNet & & & & 0.56 & 0.36 \\
%$^6$MDNet & & & & 0.63 & 0.16 \\ % VOT2014
\hline
% variations between papers of results are possible as training is done on different datasets
\end{tabular}
\\[1ex]
\parbox{\columnwidth}{\footnotesize \itshape $^1$\cite{guo-ICCV2017} \quad $^2$\cite{li-PR2018} \quad $^3$\cite{he-CVPR2018} \quad $^4$\cite{wang-CVPR2018} \quad $^5$\cite{zhang-NC2018} \quad $^6$\cite{chen-CORR2016}}
\label{tab:comparison}
\end{table}

%%%%%%%%%%%%%%%%%%%%%%%%%%%%%%%%%%%%%%%%%%%%%%%%%%%
\section{Discussion}\label{sec:discussion}
% compare and discuss the methods' details: network (branches, output, connection), training, tracker inference
This section builds on the analysis of the last section and discusses the differences of the trackers. The aim is to gain more insight into the use of Siamese networks for tracking. We will focus our discussion concerning the network function on differences in the branches, then on the layers connecting the branches and on the training. This section further discusses the differences in the tracker function.

%%%%%%%%%%%%%%%%%%%%%%%%%%%%%%%%%%%%%%%%%%%%%%%%%%%
\subsection{Network Function}
The proposed network functions either learn similarity and features jointly~\cite{fan-TNN2010, held-ECCV2016, choi-CORR2017, chen-TCSVT2017}, or assume similarity as a priori given and consider sole feature learning~\cite{tao-CVPR2016, bertinetto-VOT2016, valmadre-CVPR2017, guo-ICCV2017, wang-CORR2017}. Joint learning utilises the Siamese architecture of the network function to its full extent, while assumptions, such as a given similarity, restrict parts of the architecture.
Joint learning is currently little understood, while feature learning for the aim of compression has been extensively studied over decades by the signal processing community~\cite{nguyen-AOS2009}. Learning similarity with given features as last case is rigorously studied in statistical decision theory and machine learning.

% describe Siamese networks for tracking, common network properties to all methods (input) and initialisation, training
The attempt of all trackers is to learn a hierarchy of convolutional features of arbitrary training objects by ignoring categorisation and to train entirely offline, end-to-end the network parameters by using back-propagation and infer at runtime the object's bounding box by regressing directly \cite{held-ECCV2016} or by ranking proposed bounding box candidates given certain criteria to retrieve the best match~\cite{tao-CVPR2016} or by estimating centre position and scale in subsequent steps outside the network function~\cite{bertinetto-VOT2016, chen-TCSVT2017, valmadre-CVPR2017, fan-TNN2010, wang-CORR2017, choi-CORR2017, guo-ICCV2017}.
This approach learns very generic features which generalise to new objects and even new object categories not present in the training data~\cite{held-ECCV2016}. Siamese networks on one hand combine the expressive power of convolutional networks in the branches with real-time inference which is indispensable from an application point of view. On the other hand, the approach allows due to its simplicity a better understanding of the implications of learning jointly features and similarity.

%%%%%%%%%%%%%%%%%%%%%%%%%%%%%%%%%%%%%%%%%%%%%%%%%%%
\subsubsection{Network Branches}
% convolutional layers
All proposed trackers suggest convolutional branches inherited either from AlexNet or VGGNet with five convolutional layers. AlexNet and VGGNet allow transfer learning as these networks are extensively used in object recognition. CNNT is different as ImageNet was released 2010.
Exceptions to the five convolutional layers are YCNN with three and DCFNet with a single layer. CFNet was implemented with a different number of layers to study its effect. Valmadre et al.~\cite{valmadre-CVPR2017} report quick saturation of tracking performance with an increased depth of the branches. More than five convolutional layers yield neglectable performance gains. There is agreement that the first two convolutional layers capture local visual details such as edges and corners. These shallow features contribute most to the accurate localisation of the object. Convolutional layers three to five aggregate these details to category specific descriptions. It is argued that such semantics is important for the robustness of a tracker~\cite{danelljan-ICCVW2015}.
As agnostic tracking considers single instances of objects, the question to which extent categorial information contributes to robustness is still open. As counter example, DCFNet uses solely conv-1. DSiam uses a weighted combination of multiple layers, whereas the weights are learned during the training. In CNNT the combination is without any justification a priori given.

% max-pooling layer 
Max-pooling as it is part of AlexNet and VGGNet introduces some invariance to deformations of the object but reduces image resolution. What is an improvement to robustness is at the same time a big disadvantage. Tracking differs here significantly from categorial object recognition. This is common opinion as all trackers reduce as good trade-off the max-pooling layers to two. Chen et al.~\cite{chen-CORR2016} are the only exception who do not consider this issue.

% correlation filter
A possible alternative to max-pooling for the sake of deformations are correlation filters~\cite{valmadre-CVPR2017}. 
Such kernels offer invariant descriptions of objects and a fast convolution in the Fourier domain. Kernels can be found from data by a fast ridge regression, hence correlation filters can be applied online and even learnt online as we have seen with DCFNet. Kernels can even be computed and applied on feature maps which allows flexible use of correlation filtering and feature learning which is central to all the trackers using correlation filters such as CFNet and DCFNet. We note that max-pooling in the convolutional network and correlation filtering both contribute to invariance of deformations which might be suboptimal.

As recurrent network DCFNet propagates directly the nominator and denominator of the ridge regression back into the hidden layers while CFNet uses instead the target patch itself as memory or more precisely the moving average. We believe the latter approach looses less information as the adapted target patch contains more information than the adapted kernel which builds on the feature maps of the convolutional network.

% consequence of Pseudo Siamese networks
Another common property are the shared weights which makes the network Pseudo Siamese. The argumentation in all papers is that the sharing of weights reduces the amount of parameter by half, which helps against overfitting especially with small training datasets. What is not discussed is the effect of sharing to the connecting layers and to the outcome of the network. For example, when similarity is given by the loss function such as for SINT, it is clear that the branches are forced to learn motion invariant features which in the best case are equal on the manifold in $l_2$-space. The normalisation layer further constrains the degree of freedom of feature learning. What has clear meaning for SINT is rather unspecified for YCNN, where it is unclear what exactly is learnt in the branches and then in the connecting layers. In the most general case such as in GOTURN, branches might even learn different independent features as weight sharing is not used at all.

%%%%%%%%%%%%%%%%%%%%%%%%%%%%%%%%%%%%%%%%%%%%%%%%%%%
\subsubsection{Connection of Branches}
% three ways to connect
Research shows currently three possible ways to exploit a Siamese network for tracking. The Siamese network is either fully exploited for feature and similarity learning using fully convolutional layers without any further assumptions (CNNT, GOTURN, YCNN), or the concatenating layer is a $\star$-layer or $f_\star$-layer, i.e. similarity is defined as correlation (SiamFC, CFNet, DSiam, DCFNet), or similarity is equal to the training loss function as in the case of Two-Stream networks such as SINT where similarity is the $l_2$ distance.

% comparison, advantages
These choices have direct consequences on the tracking capabilities. For example, the $\star$-layer has clear advantages over $f_\star$ as it is fully convolutional. Features of the target patch can be correlated with the features of a search region of larger size. Two-Stream networks define similarity not as part of the network. This approach would basically allow the flexible application of different measures during training.

% consequences of assuming similarity
A given similarity gives the concatenating layers a clear meaning. The idea is then to learn in the branches the features that best match the given similarity. In some sense, both branches transform target and search patch to features that achieve highest similarity with respect to the given similarity function. Pseudo Siamese networks even learn the same feature hierarchies in both branches. Such trackers are most similar to patch matching methods using hand-crafted features~\cite{zagoruyko-CVIU2017}.

% consequence of similarity to spatiotemporal features
The concatenating fully convolutional layers build a spatiotemporal fine to coarse feature hierarchy. For example, CNNT convolutes explicitly the feature maps of target and search patch showing the object at different times. The trackers assuming similarity omit convolution or some other fusion of visual information acquired at different time instants. Our hypothesis is that these trackers neglect spatiotemporal information at all which is an important cue to reduce drift during tracking.

% translational variance
What is interesting in this respect is that although Siamese networks are able to capture spatiotemporal features, the architecture might impede the use of spatiotemporal information. For example, Fan et al.~\cite{fan-TNN2010} argue that a careful consideration of the receptive fields in the network, in their case using final four times upsampling to generate the probability map, yields a translation variant architecture, i.e. a network which exploits spatiotemporal information. A deeper analysis of GOTURN and YCNN will be needed to better understand, if their receptive fields impede or allow translational variance.

% consequence of parameter sharing
Pseudo Siamese networks are popular as the number of weights in the branches is reduced by half. It is argued that the training is more robust to overfitting, especially with small datasets. However, weight sharing has consequences to feature learning. For example, when we assume similarity, we then force the network to learn in both branches the same feature hierarchy invariant to all kinds of nuisances such as motion and appearance changes.

It is unclear what consequence weight sharing has in the case the network is able to learn spatiotemporal features. Adversarial effects might happen during training, as on the one hand the branches learn motion variant features while the fully convolutional layers compensate this effect. GOTURN avoids this effect by adapting the weights during back-propagation without such constraints. GOTURN learns a generic relationship between arbitrary motion and visual features, however, there is evidence that this relationship is dependent of the content of the target patch and the search region. Such Siamese networks might not be able to learn generic motion of a large variety of objects from examples of a few typical objects. For this, we need to explicitly model the motion, for example, by the use of multiplicative interactions~\cite{memisevic-TPAMI2013}.

%%%%%%%%%%%%%%%%%%%%%%%%%%%%%%%%%%%%%%%%%%%%%%%%%%%
\subsubsection{Network Training}
% training is crucial
The quantitative analysis shows that the selection of data and the training affects crucially the performance of the networks during inference. Training is usually done in two phases, a pre-training to transfer-learn generic spatial features of objects from labeled datasets such as ImageNet and a fine-tuning phase to adapt these features to much smaller annotated videos and to learn new spatiotemporal features.

% training loss
A second crucial factor is the loss function during back-propagation which is the standard method for training. One conclusion is that accuracy can be gained by penalising small errors~\cite{held-ECCV2016, tao-CVPR2016}. GOTURN and SINT therefore prefer $l_1$ and margin contrastive loss respectively. The former distance is however discontinuous which introduces new problems for back-propagation, while the latter loss is still $l_2$ but focuses on false sample pairs. To weight small errors larger has analogies to categorial object recognition where weighting categories with fewer instances larger is successful~\cite{lin-ICCV2017}.

As the score map is unbounded, DCFNet uses weight decay as regularisation in their $l_2$ loss. Otherwise, the optimisation is insufficiently constrained. This is a good example that a careful selection of the loss function depends on the output of the network. SiamFC and CFNet solve this problem elegantly by using a cross-entropy loss. Weight decay is usually used to prevent overfitting on small sample sizes. YCNN uses solely $l_2$ but the optimisation problem is here well constrained as probabilities are used instead of scores. 

% consequence to consistency
Estimating jointly the features in the branches and their similarity is a known problem in statistical decision theory where the experimental design and the decision function are unknown. Certain conditions on loss functions exist such that empirical risk minimisation as done by back-propagation yields Bayes consistency~\cite{nguyen-AOS2009}. Consistency of the studied trackers is currently not understood, especially when assuming the similarity function. It is for example unclear, if classes of similarity functions such as correlation interfere with particular classes of loss functions.

% translational invariance
The last section showed that a smart formulation of cross-correlation as $\star$-layer allows patch matching to a larger search region in a single feed forward step. This property has a second consequence as it introduces invariance to object translation. Translational invariance is a limiting factor to reduce drift as it impedes exploitation of spatiotemporal information. As long as the tracker reduces to pure patch matching, translational invariance has a second advantage to dataset size during training. Less data is needed as single occurrences of typical translations are sufficient due to the invariance property. 

% image context
Image context is important for recognition and tracking~\cite{grabner-CVPR2010, dinh-CVPR2011}. Neural networks have particular abilities to exploit this context. Held et al.~\cite{held-ECCV2016} show in their empirical study that VOT raw accuracy and robustness errors reduce significantly, especially in cases of occlusion, when the context is enlarged, for example by feeding the network with the whole images instead of the patches. SINT follows this approach but perhaps unaware of this insight as the motivation for SINT comes from image retrieval where processing of whole images is common.

% reinforcement learning
All trackers follow the supervised learning paradigm to train the network except Choi et al.~\cite{choi-CORR2017} who introduced reinforcement learning to tracking. RDT introduces additionally to the training loss of their matching network (YCNN) a training reward to learn a policy in choosing the target patch out of the pool. This is a direct extension to YCNN which used a confidence measure to weight target patches. The idea to choose the target patch out of a pool based on a policy learnt from data is promising, because reinforcement learning as in this work needs weakly labelled data. Furthermore, there are hints that such learning based on Q statistics considers also diversity in the pool~\cite{nebehay-MCS2013, kourosh-CVPR2018}. Episodic training with videos seems reasonable. Such policies could be extended to other building blocks of tracking, such as $f_\text{box}$ and the scaling of bounding boxes.

% control of training data
There is common agreement to use ImageNet-12, 14, 15 and ALOV for training and VOT and OTB for evaluation. For the sake of a better evaluation, the use of OTB and VOT for training as done for YCNN and RDT should be avoided. To guarantee a fair comparison, proprietary training data should also be avoided. The quantitative analysis suggests that training i.e. training data and training procedures need more control for a reasonable evaluation.

%%%%%%%%%%%%%%%%%%%%%%%%%%%%%%%%%%%%%%%%%%%%%%%%%%%
\subsection{Tracker Function}
% composition of the tracker
The tracker functions are composed basically of two partial functions. A partial tracker function $\delta_0(\cdot)$ that prepares the sequence of images and past measurements and a function $\delta_1(\cdot)$ that generates the bounding box by using the network function $\sigma(\cdot)$. This pre- and post-processing is kept rather simple by using heuristics.

% discuss class and measurements of tracker functions
The previous section analysed that SINT and SiamFC are Type-1 trackers, i.e. both consider the single initial target patch, CNNT, GOTURN and DCFNet are of Type-2 Markov and CFNet, YCNN and RDT are full Type-2 trackers. CFNet adapts steadily the previous target patches by a moving average. YCNN and RDT loose a target patch from their finite pool in case of low confidence or after a constant number of time steps. These trackers might loose the initial target patch which we see disadvantageous for a Type-2 tracker. Such approaches are prone to drift, although adapting the target patch improves sometimes accuracy and robustness~\cite{held-ECCV2016}.

% delta_0
Looking closer at $\delta_0(\cdot)$, trackers using search regions mostly crop them adhoc by a certain multiple of the bounding box which assumes some constraints on object and camera motion. SINT is here an exception as it uses instead region CNNs in both branches.

% delta_1
Usually, $\delta_1(\cdot)$ applies greedy optimisation by using several scaled versions of the target patch to $\sigma(\cdot)$, where the patch with best fit, either highest correlation or smallest error, determines the bounding box. RDT uses also patch translations, but it is unclear how much improvement can be gained. DSiam scales instead of the target patch the search region. Only SINT, GOTURN and CNNT consider $f_\text{box}$ as part of $\sigma(\cdot)$ with the clear advantage that $f_\text{box}$ is approximated by $\sigma(\cdot)$. While GOTURN regresses directly the bounding box, SINT uses region proposals by prior sampling of candidate boxes in the search region. This evaluation of proposals is done in a single feed forward step, however, identification of the best fit needs explicit comparison outside of the network. CNNT goes here an interesting new way by regressing the dual of the bounding box i.e. the four corners that determine a bounding box.

% pool of patches
As YCNN and RDT maintain a pool of target patches, YCNN uses a weighted average based on confidence to compute the bounding box. We see this idea sceptical as the average is sensitive to even small errors. RDT develops this idea further and uses a policy function to choose from this pool. We see RDT's policy function as clear improvement over YCNN's confidence weighted average.

% new measurements
All trackers consider the measurement $Y_t$ as bounding box at time $t$. DSiam goes here an interesting different way as DSiam maintains additionally the affine transformations $U$ and $V$. $U$ compensates appearance changes caused by object deformation. This transformation can be seen as some form of state prediction which is well known in Bayesian tracking~\cite{challa-book2011}. $V$ suppresses background features similar to the appearance features of the object. This transformation makes tracking less vulnerable to drift which is an idea similar to background suppression in mean shift tracking~\cite{comaniciu-TPAMI2002}.

% reccurent
Siamese networks for Type-2 tracking can be seen as simplest unrolled recurrent network. Most papers except Bertinetto et al.~\cite{bertinetto-VOT2016} and Wang et al.~\cite{wang-CORR2017} do not mention this relationship. In most cases the measurements $Y_t$ act as memory, for example the moving average in CFNet. DCFNet goes here a different way by directly encoding the online ridge regression into the network by using feedback loops in the hidden layers.

%%%%%%%%%%%%%%%%%%%%%%%%%%%%%%%%%%%%%%%%%%%%%%%%%%%
\section{Conclusion and Future Work}\label{sec:conclusion}
% conclude the paper
This study compares the results and the design of nine recent trackers. Rather than giving a broad overview of trackers using neural networks for inference and learning, this work focuses on trackers using Siamese neural networks. On the one hand this approach allows a deeper analysis of the tracker details, on the other hand Siamese networks are an excellent starting point to study neural networks for tracking as they are the simplest networks for matching (correspondence, association) problems. The literature on deep learning for tracking is incredibly growing since 2015, so it is very difficult to give an elaborate survey about all the literature available. Siamese networks integrate efficiently feature extraction and the temporal matching in the simplest way. They have so far shown state-of-the-art performance in accuracy and robustness. Instead of proposing a new dataset, this work identified appropriate and comparable data from the existing literature. Quantitative results of the nine trackers from this existing papers and benchmarks are compared with the conclusion that the current methodology shows problems with the reproducibility and the comparability of results.
This study also compares the branches of the Siamese networks, their layers connecting these branches, and the whole tracker functions themselves. For this study, we introduce a novel Lisp-like formalism which allows to recognise even tiny details much better than for example a graphical comparison or a descriptive comparison. This assumes a certain functional design of trackers as tracker functions. A foundation for this tracker design is given by a formulation of the problem based on the theory of machine learning and by the interpretation of a tracker as a decision function.

% consequences of the connectionist view
Should we now start re-implementing various traditional trackers as networks? We should definitely build upon the existing knowledge of traditional tracking. The traditional view of tracking suggests a variety of building blocks such as the extraction of features, matching, localisation and the temporal update of the object description. Nearly all considered neural trackers study the matching and feature tasks by integrating both tasks into a single Siamese network. Then DSiam with background suppression similar to Mean-shift~\cite{comaniciu-TPAMI2002} is another good example.
But as we have seen the advantage of neural networks is the ability to integrate elegantly various different tasks and heterogeneous computational processes into a single network~\cite{doersch-ICCV2017, ranftl-DAGM2014}, very different to the limited traditional view of designing the tasks of tracking sequentially. Whole trackers are then trainable end-to-end on given videos by using back-propagation, as long as the functional tasks are differentiable. This goes certainly beyond the capability of traditional trackers. Novel networks might even integrate online learning as complementary task.

% initialisation
Current work has not even begun to understand the potential of neural networks to tackle the current challenges of tracking (Section~\ref{sec:challenges}). More complex networks beyond Siamese will add further important tasks to the network. For example, memories such as Long Short-Term Memory (LSTM)~\cite{hochreiter-NC1997} or attentional mechanisms~\cite{bazzani-ICML2011, graves-nature2016, kahou-CVPRW2017, kosiorek-NIPS2017} together with saliency features~\cite{kuemmerer-ICCV2017} might offer a future solution to the problem of initialisation. This research is interdisciplinary and will trigger mutual insights in computer vision and cognitive psychology~\cite{grossberg-TCS2000, valuch-JV2017, marcus-BOOK2003}.
Another line of research is the combination of detection and tracking~\cite{tao-CVPR2016, feichtenhofer-ICCV2017, luo-CVPR2018} as joint task. Both tasks are to a certain extent complimentary. We have seen in in the analysis of CNNT that tracking is motion variant while detection in images is a task invariant to object motion~\cite{fan-TNN2010}.

% complexity
Memory might also be the key to tackle varying complexity as memory allows the network to adapt over time the description of objects. But this assumes still a better understanding of Siamese networks, especially the joint learning of features and similarity~\cite{nguyen-ICML2004} in the spatiotemporal domain. The ideas of multi-layer features and unification of loss and similarity function as shown by Two-Stream networks need further studies. Geometrical transformations should be explicitly considered to reduce the complexity, for example by using transformer~\cite{jaderberg-NIPS2015} or pose networks~\cite{rad-CVPR2018} pre-trained on artificial data as part of the network function. Attentional mechanisms for adaptive feature selection are also showing promising results to adapt complexity in the description~\cite{lukezic-CVPR2017, he-CVPR2018, wang-CVPR2018, zhu-CVPR2018}.
Triplet networks are the logical continuation of the Siamese approach~\cite{hoffer-CORR2014, wang-CVPR2014}. Content-free spatiotemporal descriptions are needed as such descriptions might be learnable on a few typical objects and then be applicable to a variety of unknown objects. Generative models~\cite{jepson-TPAMI2003} and networks~\cite{kosiorek-CORR2018} and the idea of multiplicative triplets~\cite{memisevic-TPAMI2013} are a good starting point for potential solutions and would also incorporate the traditional task of prediction more elaborate than DSiam. 
Another limitation to accuracy and robustness is the use of the axis aligned bounding box as measurement. Further research between rotating bounding boxes and pixelwise segmentations as well as parts-based measurements is needed~\cite{godec-ICCV2011, nebehay-CVPR2015, liu-CVPR2015, cheng-CVPR2018}. Parts-based descriptions have the advantage of being more robust to partial occlusions due to the redundancy of parts for estimating the latent variables~\cite{gao-TIP2018}.

% uncertainty
Issues such as uncertainty are currently neglected, although uncertainty has always been a very important topic for tracking. For example, filtering and sequential Monte Carlo methods have been extensively used for tracking in the past~\cite{lei-TSMC2008, isard-IJCV1998, khan-TPAMI2006}. Formulating the tracker function as posterior and using Bayesian analysis is well known~\cite{challa-book2011}. It would be promising to study the network under this Bayesian view, i.e. interpreting the network function as likelihood function. For current network functions, i.e. frequentistic functions, it is interesting to study confidence intervals, consistency of learning~\cite{nguyen-AOS2009}.

% comparison
Current work is unfortunately not reproducing the results of their peers. This is a severe problem as it affects the scientific methodology of empirical analysis in its roots. This study identifies the different training data and the annual change of evaluation data such as for VOT as one of the main reasons. The community has now a common understanding to use OTB and VOT for evaluation and ALOV and ImageNet for training. But as VOT is further evolving and new datasets are created, the situation will not improve.
One solution is a common agreement and standard for training similar to the activities in evaluation. A community driven choice of such datasets, for example OTB-2013 or VOT-2015, should kept permanent and available over decades. Measures such as AuC, raw A/R-scores should become standard to guarantee a reasonable comparison of results among the work and papers, respectively.
A new formalism for qualitative comparison as suggested in this work by the introduction of a clear functional formalism based on prefix notation is further needed. Such comparison is essential to identify common design patterns for a larger number of trackers.

%%%%%%%%%%%%%%%%%%%%%%%%%%%%%%%%%%%%%%%%%%%%%%%%%%%
\section*{Acknowledgments}
I thank my wife and my daughter for the hours writing this manuscript and Andreas Kriechbaum-Zambini for his patience. I thank my colleagues Axel Wei{\ss}enfeld and Gustavo Fern\'andez for their valuable comments. This research has received funding from the EU ARTEMIS Joint Undertaking under grant agreements no. 621429 (EMC{$^2$}) and from the FFG (Austrian Research Promotion Agency) on behalf of BMVIT, The Federal Ministry of Transport, Innovation and Technology. This work was supported by the AIT strategic research programme Intelligent Cameras and Video Analytics 2017-18.
\ifCLASSOPTIONcaptionsoff
  \newpage
\fi
\bibliographystyle{IEEEtran}
\bibliography{IEEEabrv,survey-min}
\begin{IEEEbiography}[{\includegraphics[width=1in,height=1.25in,clip,keepaspectratio]{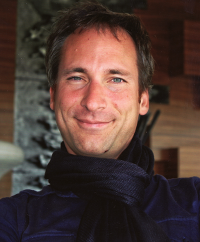}}]{Roman Pflugfelder} is Scientist at the AIT Austrian Institute of Technology and lecturer at TU Wien. He received in 2002 a MSc degree in informatics at TU Wien and in 2008 a PhD in telematics at the TU Graz, Austria. In 2001, he was academic visitor at the Queensland University of Technology, Australia. His research focuses on visual motion analysis, tracking and recognition applied to automated video surveillance. He aims to combine sciences and theories in novel ways to gain theoretical insights into learning and inference in complex dynamical systems and to develop practical algorithms and computing systems. Roman contributed with more than 55 papers and patents to research fields such as camera calibration, object detection, object tracking, event recognition where he received awareness of media as well as several awards and grants for his scientific achievements. Roman is senior project leader at AIT where he has been managing cooperations among universities, companies and governmental institutions. Roman co-organised the Visual Object Tracking Challenges VOT'13-14 and VOT'16-18 and was program co-chair of AVSS'15. Currently he is steering committee member of AVSS. He is regular reviewer for major computer vision conferences and journals.
\end{IEEEbiography}

\end{document}